\documentclass{article} % For LaTeX2e
\usepackage{iclr2024_conference,times}

\usepackage{amsmath,amsfonts,bm}

\def\eqref#1{equation~\ref{#1}}
\def\1{\bm{1}}

\DeclareMathAlphabet{\mathsfit}{\encodingdefault}{\sfdefault}{m}{sl}
\SetMathAlphabet{\mathsfit}{bold}{\encodingdefault}{\sfdefault}{bx}{n}

\usepackage{hyperref}
\usepackage{url}

\usepackage{hyperref}
\usepackage{url}
\usepackage{amsmath}
\usepackage{mathabx}
\usepackage{bm}
\usepackage{algorithmicx}
\usepackage{algorithm}
\usepackage{algpseudocode}
\usepackage{bbm}
\usepackage{graphicx}
\usepackage{subcaption}
\usepackage{wrapfig}
\usepackage[framemethod=TikZ]{mdframed}
\usepackage[most]{tcolorbox}

\usepackage{amsthm}
\usepackage{cleveref}
\theoremstyle{plain}
\newtheorem{theorem}{Theorem}[section]

\newtheorem{proposition}{Proposition}[section]

\theoremstyle{definition}

\newtheorem{definition}{Definition}[section]

\theoremstyle{remark}
\newtheorem{remark}{Remark}[section]

\mdfsetup{
	middlelinecolor =  none,
	middlelinewidth = 1pt,
	backgroundcolor = blue!10,
	roundcorner = 5pt,
}

\newcommand{\parag}[1]{\textbf{#1}}
\newcommand{\highlight}[1]{\colorbox{blue!10}{\ensuremath{#1}}}

\title{Soft Mixture Denoising: Beyond the Expressive Bottleneck of Diffusion Models}

\author{Yangming Li, Boris van Breugel, Mihaela van der Schaar \\
Department of Applied Mathematics and Theoretical Physics \\
University of Cambridge \\
\texttt{\{yl874,bv292,mv472\}}@cam.ac.uk
}

\iclrfinalcopy % Uncomment for camera-ready version, but NOT for submission.
\begin{document}

\maketitle

\begin{abstract}
Because diffusion models have shown impressive performances in a number of tasks, such as image synthesis, there is a trend in recent works to prove (with certain assumptions) that these models have strong approximation capabilities. In this paper, we show that current diffusion models actually have an \textit{expressive bottleneck} in backward denoising and some assumption made by existing theoretical guarantees is too strong. Based on this finding, we prove that diffusion models have unbounded errors in both local and global denoising. In light of our theoretical studies, we introduce \textit{soft mixture denoising} (SMD), an expressive and efficient model for backward denoising. SMD not only permits diffusion models to well approximate any Gaussian mixture distributions in theory, but also is simple and efficient for implementation. Our experiments on multiple image datasets show that SMD significantly improves different types of diffusion models (e.g., DDPM), espeically in the situation of few backward iterations.
\end{abstract}

\section{Introduction}

	Diffusion models (DMs)~\citep{sohl2015deep} have become highly popular generative
	models for their impressive performance in many research domains---including high-resolution image synthesis~\citep{NEURIPS2021_49ad23d1}, natural language generation~\citep{li2022diffusionlm}, speech processing~\citep{kong2021diffwave}, and medical image analysis~\citep{pinaya2022brain}. 
	
	\parag{Current strong approximator theorems.} To explain the effectiveness of diffusion models, recent work~\citep{lee2022convergence,lee2022polynomial,chen2023sampling} provided theoretical guarantees (with certain assumptions) to show that diffusion models can approximate a rich family of data distributions with arbitrarily small errors. For example, \citet{chen2023sampling} proved that the generated samples from diffusion models converge (in distribution) to the real data under ideal conditions. Since it is generally intractable to analyze the non-convex optimization of neural networks, a potential weakness of these works is that they all supposed \textit{bounded score estimation errors}, which means the prediction errors of denoising functions (i.e., reparameterized score functions) are bounded.
	
	\parag{Our limited approximation theorems.} In this work, we take a first step towards the opposite direction: Instead of explaining why diffusion models are highly effective, we show that their approximation capabilities are in fact limited and the assumption of \textit{bounded score estimation errors} (made by existing theoretical guarantees) is too strong. 
	
	In particular, we show that current diffusion models suffer from an \textbf{expressive bottleneck}---the Gaussian parameterization of backward probability $p_{\theta}(\mathbf{x}_{t-1} \mid \mathbf{x}_t)$ is not expressive enough to fit the (possibly multimodal) posterior probability $q(\mathbf{x}_{t-1} \mid \mathbf{x}_t)$. Following this, we prove that \textit{diffusion models have arbitrarily large denoising errors for approximating some common data distributions $q(\mathbf{x}_0)$ (e.g., Gaussian mixture)}, which indicates that some assumption of prior works---bounded score estimation errors---is too strong, which undermines their theoretical guarantees. Lastly and importantly, we prove that \textit{diffusion models will have an arbitrarily large error in matching the learnable backward process $p_{\theta}(\mathbf{x}_{0:T})$ with the predefined forward process $q(\mathbf{x}_{0:T})$}, even though matching these is the very optimization objective of current diffusion models~\citep{ho2020denoising,song2021scorebased}. This finding indicates that diffusion models might fail to fit complex data distributions.
	
	\begin{figure}[htb]
		\centering
		\begin{subfigure}{0.48\textwidth}
			\includegraphics[width=\linewidth]{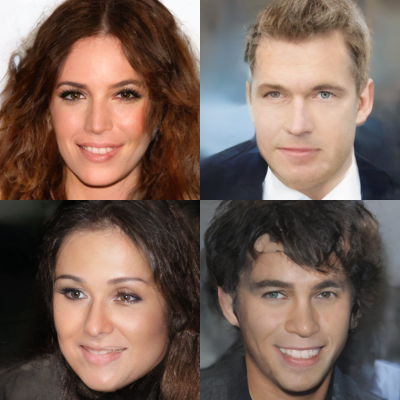}
			\caption{Baseline: vanilla LDM; FID: $11.29$.}
		\end{subfigure}
		\hfill
		\begin{subfigure}{0.48\textwidth}
			\includegraphics[width=\linewidth]{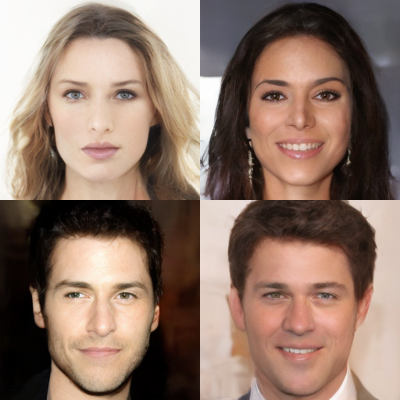}
			\caption{Our model: LDM w/ SMD; FID: $6.85$.}
		\end{subfigure}
		\caption{\textbf{SMD improves quality and reduces the number of backward iterations.} Results for CelebA-HQ $256 \times 256$ with \textit{only $100$ backward iterations}, for LDM with and without SDM. SDM achieves better realism and FID. Achieving the same FID with vanilla LDM would require $8\times$ more steps (see Fig. \ref{fig:few iters}). Note that SMD differs from fast samplers (e.g., DDIM~\citep{song2021denoising} and DPM~\citep{lu2022dpm}): \textit{while those methods focus on deterministic sampling and numerical stability, SMD improves the expressiveness of diffusion models}.}
		\label{fig:demo}
	\end{figure}

	\parag{Our method: Soft Mixture Denoising (SMD).} In light of our theoretical findings, we propose Soft Mixture Denoising (SMD), which aims to represent the hidden mixture components of the posterior probability with a continuous relaxation. We prove that \textit{SMD permits diffusion models to accurately approximate any Gaussian mixture distributions}. For efficiency, we reparameterize SMD and derive an upper bound of the negative log-likelihood for optimization. All in all, this provides a new backward denoising paradigm to the diffusion models that improves expressiveness and permits few backward iterations, yet retains tractability.

	\parag{Contributions.} In summary, our contributions are threefold:
	\begin{enumerate}\itemsep0em 
		\item In terms of theory, we find that current diffusion models suffer from an \textit{expressive bottleneck}. We prove that the models have unbounded errors in both local and global denoising, demonstrating that the assumption of \textit{bounded score estimation errors} made by current theoretical guarantees is too strong; 
		\item In terms of methodology, we introduce SMD, an expressive backward denoising model. Not only does SMD permit the diffusion models to accurately fit Gaussian mixture distributions, but it is also simple and efficient to implement;
		\item In terms of experiments, we show that SMD significantly improves the generation quality of different diffusion models (DDPM~\citep{ho2020denoising}, DDIM~\citep{song2021denoising}, ADM~\citep{NEURIPS2021_49ad23d1}, and LDM~\citep{rombach2022high}), especially for few backward iterations---see Fig.~\ref{fig:demo} for a preview. Since SMD lets diffusion models achieve competitive performances at a smaller number of denoising steps, it can speed up sampling and reduce the cost of existing models.
	\end{enumerate}

\section{Background: Discrete-time Diffusion Models}

	In this section, we briefly review the mainstream architecture of diffusion models in discrete time (e.g., DDPM~\citep{ho2020denoising}). The notations and terminologies introduced below are necessary preparations for diving into subsequent sections.
	
	A diffusion model typically consists of two Markov chains of $T$ steps. One of them is the forward process---also known as the diffusion process---which incrementally adds Gaussian noises to the real sample $\mathbf{x}_0 \in \mathbb{R}^{D}, D \in \mathbb{N}$, giving a chain of variables $\mathbf{x}_{1:T} = [\mathbf{x}_1, \mathbf{x}_2, \cdots, \mathbf{x}_T]$:
	\begin{equation}
		\label{eq:forward def}
		q(\mathbf{x}_{1:T} \mid \mathbf{x}_0) = \prod_{t=1}^T q(\mathbf{x}_t \mid \mathbf{x}_{t-1}), \ \ \ q(\mathbf{x}_t \mid \mathbf{x}_{t-1}) = \mathcal{N}(\mathbf{x}_t; \sqrt{1 - \beta_t} \mathbf{x}_{t-1}, \beta_t \mathbf{I}),
	\end{equation}
	where $\mathcal{N}$ denotes a Gaussian distribution, $\mathbf{I}$ represents an identity matrix, and $\beta_t, 1 \le t \le T$ are a predefined variance schedule. By properly defining the variance schedule,  the last variable $\mathbf{x}_T$ will approximately follow a normal Gaussian distribution.

	The second part of diffusion models is the \textit{backward} (or \textit{reverse}) \textit{process}. Specifically speaking, the process first draws an initial sample $\mathbf{x}_T$ from a standard Gaussian $p(\mathbf{x}_T) = \mathcal{N}(\mathbf{0}, \mathbf{I})$ and then gradually denoises it into a sequence of variables $\mathbf{x}_{T-1:0} = [\mathbf{x}_{T-1}, \mathbf{x}_{T-2}, \cdots, \mathbf{x}_0]$:
	\begin{equation}
		\label{eq:backward proc}  
		p_{\theta}(\mathbf{x}_{T:0}) = p(\mathbf{x}_T) \prod_{t=T}^1 p_{\theta}(\mathbf{x}_{t-1} \mid \mathbf{x}_{t}), \ \ \ p_{\theta}(\mathbf{x}_{t-1} \mid \mathbf{x}_{t}) = \mathcal{N}(\mathbf{x}_{t-1}; \bm{\mu}_{\theta}(\mathbf{x}_t, t), \sigma_t\mathbf{I}),
	\end{equation}
	where $\sigma_t \mathbf{I}$ is a predefined covariance matrix and $\bm{\mu}_\theta$ is a learnable module with the parameter $\theta$ to predict the mean vector. Ideally, the learnable backward probability $p_{\theta}(\mathbf{x}_{t-1} \mid \mathbf{x}_{t})$ is equal to the inverse forward probability $q(\mathbf{x}_{t-1} \mid \mathbf{x}_t)$ at every iteration $t \in [1, T]$ such that the backward process is just a reverse version of the forward process.

	Since the exact negative log-likelihood $\mathbb{E}[-\log p_{\theta}(\mathbf{x}_0)]$ is computationally intractable, common practices adopt its upper bound $\mathcal{L}$ as the loss function
	\begin{equation}
		\label{eq:jensen}
		\begin{aligned}
			\mathbb{E}_{\mathbf{x}_0 \sim q(\mathbf{x}_0)}[-\log p_{\theta}(\mathbf{x}_0)] & \le \underbrace{\mathbb{E}_q [\mathcal{D}_{\mathrm{KL}}[q(\mathbf{x}_T \mid \mathbf{x}_0), p(\mathbf{x}_T) ]]}_{\mathcal{L}_T}  + \underbrace{\mathbb{E}_q [-\log p_{\theta} (\mathbf{x}_0 \mid \mathbf{x}_1)  ]}_{\mathcal{L}_0} \\
			& + \sum_{1 < t \le T} \underbrace{\mathbb{E}_q [ \mathcal{D}_{\mathrm{KL}} [ q(\mathbf{x}_{t-1} \mid \mathbf{x}_t, \mathbf{x}_0), p_{\theta} (\mathbf{x}_{t-1} \mid \mathbf{x}_{t})  ]]}_{\mathcal{L}_{t-1}} = \mathcal{L},
		\end{aligned}
	\end{equation}	
	where $\mathcal{D}_{\mathrm{KL}}$ denotes the KL divergence. Every term of this loss has an analytic form so that it is computationally optimizable. \cite{ho2020denoising} further applied some reparameterization tricks to the loss $\mathcal{L}$ for reducing its variance. As a result, the module $\bm\mu_{\theta}$ is reparameterized as
	\begin{equation}
		\label{eq:mean def}
		\bm{\mu}_{\theta}(\mathbf{x}_t, t) = \frac{1}{\sqrt{\alpha_t}} \Big(\mathbf{x}_t - \frac{\beta_t}{\sqrt{1 - \widebar{\alpha}_t}}\bm{\epsilon}_{\theta}(\mathbf{x}_t, t) \Big),
	\end{equation}
	where $\alpha_t = 1 - \beta_t$, $\widebar{\alpha}_t = \prod_{t'=1}^t \alpha_{t'}$, and $\bm{\epsilon}_{\theta}$ is parameterized by neural networks. Under this popular scheme, the loss $\mathcal{L}$ is finally simplified as
	\begin{equation}
		\label{eq:simplified loss}
		\mathcal{L} = \sum_{t=1}^T \mathbb{E}_{\mathbf{x}_0 \sim q(\mathbf{x}_0), \bm{\epsilon} \sim \mathcal{N}(\mathbf{0}, \mathbf{I})} \Big[ \| \bm{\epsilon} - \bm{\epsilon}_{\theta} (\sqrt{\widebar{\alpha}_t} \mathbf{x}_0 + \sqrt{1 - \widebar{\alpha}_t} \bm{\epsilon}, t) \|^2 \Big],
	\end{equation}
	where the denoising function $\bm{\epsilon}_{\theta}$ is tasked to fit Gaussian nosie $\bm{\epsilon}$.

\section{Theory: DMs Suffer from an Expressive Bottleneck}
\label{sec:theory}

	In this section, we first show that the Gaussian denoising paradigm leads to an \textit{expressive bottleneck} for diffusion models to fit multimodal data distribution $q(\mathbf{x}_0)$. Then, we properly define two errors $\mathcal{M}_t, \mathcal{E}$ that measure the approximation capability of general diffusion models and prove that they can both be unbounded for current models.
	
\subsection{Limited Gaussian Denoising}

	The core of diffusion models is to let the learnable backward probability $p_{\theta}(\mathbf{x}_{t-1} \mid \mathbf{x}_t)$ at every iteration $t$ fit the posterior forward probability $q(\mathbf{x}_{t-1} \mid \mathbf{x}_t)$. From Eq.~(\ref{eq:backward proc}), we see that the learnable probability is configured as a simple Gaussian $\mathcal{N}(\mathbf{x}_{t-1}; \bm{\mu}_{\theta}(\mathbf{x}_t, t), \sigma_t\mathbf{I})$. While this setup is analytically tractable and computationally efficient, our proposition below shows that its approximation goal $q(\mathbf{x}_{t-1} \mid \mathbf{x}_t)$ might be much more complex.

	\begin{proposition}[Non-Gaussian Inverse Probability]
	\label{lemma:posterior form}
	
	For the diffusion process defined in Eq.~(\ref{eq:forward def}), suppose that the real data follow a Gaussian mixture: $q(\mathbf{x}_0) = \sum_{k=1}^K w_k \mathcal{N}(\mathbf{x}_0; \bm{\mu}_k, \bm{\Sigma}_k)$, which consists of $K$ Gaussian components with mixture weight $w_k$, mean vector $\bm{\mu}_k$, and covariance matrix $\bm{\Sigma}_k$, then the posterior forward probability $q(\mathbf{x}_{t-1} \mid \mathbf{x}_t)$ at every iteration $t \in [1, T]$ is another mixture of Gaussian distributions:
    \begin{equation}
        q(\mathbf{x}_{t-1} \mid \mathbf{x}_{t}) = \sum_{k=1}^K w_k' \mathcal{N}  ( \mathbf{x}_{t-1}; \bm{\mu}_k',  \bm{\Sigma}_k'),
    \end{equation}
    where $w_k', \bm{\mu}_k'$ depend on both variable $\mathbf{x}_t$ and $\bm{\mu}_t$.
    
	 \begin{remark}
		The Gaussian mixture in theory is a universal approximator of smooth probability densities~\citep{dalal1983approximating,goodfellow2016deep}. Therefore, this proposition implies that the posterior forward probability $q(\mathbf{x}_{t-1} \mid \mathbf{x}_t)$ can be arbitrarily complex.
	\end{remark}
	
	\begin{proof}
		The proof to this proposition is fully provided in Appendix~\ref{appendix:proof to posterior}.
	\end{proof}
	
	\end{proposition}

	While diffusion models perform well in practice, we can infer from above that the Gaussian denoising paradigm $p_{\theta}(\mathbf{x}_{t-1} \mid \mathbf{x}_t) = \mathcal{N}(\mathbf{x}_{t-1}; \bm{\mu}_{\theta}(\mathbf{x}_t, t), \sigma_t\mathbf{I})$ causes a bottleneck for the backward probability to fit the potentially multimodal distribution $q(\mathbf{x}_{t-1} \mid \mathbf{x}_t)$. Importantly, this problem is not rare since real-world data distributions are commonly non-Gaussian and multimodal. For example, classes in a typical image dataset are likely to form separate modes, and possibly even multiple modes per class (e.g. different dog breeds).
	
	\begin{mdframed}[leftmargin=0pt, rightmargin=0pt, innerleftmargin=10pt, innerrightmargin=10pt, skipbelow=0pt]
		\textbf{\emph{Takeaway}}: The posterior forward probability $q(\mathbf{x}_{t-1} \mid \mathbf{x}_t)$ can be arbitrarily complex for the Gaussian backward probability $p_{\theta}(\mathbf{x}_{t-1} \mid \mathbf{x}_{t}) = \mathcal{N}(\mathbf{x}_{t-1}; \bm{\mu}_{\theta}(\mathbf{x}_t, t), \sigma_t\mathbf{I})$ to approximate. We call this problem the \textit{expressive bottleneck} of diffusion models.
	\end{mdframed}

\subsection{Denoising and Approximation Errors}

	To quantify the impact of this expressive bottleneck, we define two error measures in terms of local and global denoising errors, i.e., the discrepancy between forward process $q(\mathbf{x}_{0:T})$ and backward process $p_{\theta}(\mathbf{x}_{0:T})$.
	
	\parag{Derivation of the local denoising error.} Considering the form of loss term $\mathcal{L}_{t-1}$ in Eq.~(\ref{eq:jensen}), we apply the KL divergence to estimate the approximation error of every learnable backward probability $p_{\theta}(\mathbf{x}_{t-1} \mid \mathbf{x}_t), t \in [1, T]$ to its reference $q(\mathbf{x}_{t-1} \mid \mathbf{x}_t)$ as $\mathcal{D}_{\mathrm{KL}}[q(\mathbf{x}_{t-1} \mid \mathbf{x}_t),  p_{\theta}(\mathbf{x}_{t-1} \mid \mathbf{x}_t)]$. Since the error depends on variable $\mathbf{x}_t$, we normalize it with density $q(\mathbf{x}_t)$ into $\mathbb{E}[\mathcal{D}_{\mathrm{KL}}[\cdot]] = \int_{\mathbf{x}_t} q(\mathbf{x}_t) \mathcal{D}_{\mathrm{KL}}[\cdot] d\mathbf{x}_t$. Importantly, we take the infimum of this error over the parameter space $\Theta$ as $\inf_{\theta \in \Theta} ( \int_{\mathbf{x}_t} q(\mathbf{x}_t) \mathcal{D}_{\mathrm{KL}}[q(\cdot), p_{\theta}(\cdot)] d\mathbf{x}_t )$, which means neural networks are globally optimized. In light of the above derivation, we have the following definition.
	\begin{definition}[Local Denoising Error]
		
		For every learnable backward probability $p_{\theta}(\mathbf{x}_{t-1} \mid \mathbf{x}_t), 1 \le t \le T$ in a diffusion model, its error of best approximation (i.e., parameter $\theta$ is globally optimized) to the reference $q(\mathbf{x}_{t-1} \mid \mathbf{x}_t)$ is defined as
		\begin{equation}
			\label{eq:def of error metric}
			\begin{aligned}
				\mathcal{M}_t & = \inf_{\theta \in \Theta}\Big(\mathbb{E}_{\mathbf{x}_t \sim q(\mathbf{x}_t)}[\mathcal{D}_{\mathrm{KL}}[q(\mathbf{x}_{t-1} \mid \mathbf{x}_t),  p_{\theta}(\mathbf{x}_{t-1} \mid \mathbf{x}_t)]]\Big) \\ & = \inf_{\theta \in \Theta} \Big( \int_{\mathbf{x}_t} \underbrace{q(\mathbf{x}_t)}_{\textrm{Density Weight}} \underbrace{\mathcal{D}_{\mathrm{KL}}[q(\mathbf{x}_{t-1} \mid \mathbf{x}_t),  p_{\theta}(\mathbf{x}_{t-1} \mid \mathbf{x}_t)]}_{\textrm{Denoising Error w.r.t. the Input}~\mathbf{x}_t}  d\mathbf{x}_t \Big),
			\end{aligned}
		\end{equation}		
		where space $\Theta$ represents the set of all possible parameters. Note that the inequality $\mathcal{M}_t \ge 0$ always holds because KL divergence is non-negative.
		
	\end{definition}

	\parag{Significance of the global denoising error.} Current practices~\citep{ho2020denoising} expect the backward process $p_{\theta}(\mathbf{x}_{0:T})$ to exactly match the forward process $q(\mathbf{x}_{0:T})$ such that their marginals at iteration $0$ are equal: $q(\mathbf{x}_0) = p_{\theta}(\mathbf{x}_0)$. For example, \citet{song2021scorebased} directly configured the backward process as the reverse-time diffusion equation. Hence, we have the following error definition to measure the global denoising capability of diffusion models.
	
	\begin{definition}[Global Denoising Error]
		The discrepancy between learnable backward process $p_{\theta}(\mathbf{x}_{0:T})$ and predefined forward process $q(\mathbf{x}_{0:T})$ is estimated as
		\begin{equation}
			\mathcal{E} = \inf_{\theta \in \Theta} \Big( \mathcal{D}_{\mathrm{KL}}[q(\mathbf{x}_{0:T}), p_{\theta}(\mathbf{x}_{0:T})] \Big),
		\end{equation}
		where again $\mathcal{E} \ge 0$ always holds since KL divergence is non-negative.
		
	\end{definition}

\subsection{Limited Approximation Theorems}

	In this part, we prove that the above defined errors are unbounded for current diffusion models.\footnote{It is also worth noting that these errors already overestimate the performances of diffusion models, since their definitions involve an infimum operation $\inf_{\theta \in \Theta}$.}
	\begin{theorem}[Uniformly Unbounded Denoising Error]
	\label{prop:upper bound}
		
		For the diffusion process defined in Eq.~(\ref{eq:forward def}) and the Gaussian denoising process defined in Eq.~(\ref{eq:backward proc}), there exists a continuous data distribution $q(\mathbf{x}_0)$ (more specifically, Gaussian mixture) such that $\mathcal{M}_t$ is uniformly unbounded---given any real number $N\in\mathbb{R}$, the inequality $\mathcal{M}_t > N$ holds for every denoising iteration $t \in [1, T]$.
	
		\begin{proof}
			We provide a complete proof to this theorem in Appendix \ref{appendix:upper bound}.
		\end{proof}
		
	\end{theorem}
	The above theorem not only implies that current diffusion models fail to fit some multimodal data distribution $q(\mathbf{x}_t)$ because of their limited expressiveness in local denoising, but also indicates that the assumption of \textit{bounded score estimation errors} (i.e., bounded denoising errors) is too strong. Consequently, this undermines existing theoretical guarantees~\citep{lee2022convergence,chen2023sampling} that aim to prove that diffusion models are universal approximates.
	
	\begin{mdframed}[leftmargin=0pt, rightmargin=0pt, innerleftmargin=10pt, innerrightmargin=10pt, skipbelow=0pt]
		\textbf{\emph{Takeaway}}: The denoising error $\mathcal{M}_t$ of current diffusion models can be arbitrarily large at every denoising step $t \in [1, T]$. Thus, the assumption of \textit{bounded score estimation errors} made by existing theoretical guarantees is too strong.
	\end{mdframed}
	
	Based on Theorem~\ref{prop:upper bound} and Proposition~\ref{lemma:posterior form}, we finally show that the global denoising error $\mathcal{E}$ of current diffusion models is also unbounded.
	
	\begin{theorem}[Unbounded Approximation Error]
		\label{theorem:inconsistent backward proc}
		
		For the forward and backward processes respectively defined in Eq.~(\ref{eq:forward def}) and Eq.~(\ref{eq:backward proc}), given any real number $N\in \mathbb{R}$, there exists a continuous data distribution $q(\mathbf{x}_0)$ (specifically, Gaussian mixture) such that $\mathcal{E} > N$.
		\begin{proof}
			A complete proof to this theorem is offered in Appendix~\ref{proof:inconsistent backward proc}.
		\end{proof}
		
	\end{theorem}
	Since the negative likelihood $\mathbb{E}[-\log p_{\theta}(\mathbf{x}_0)]$ is computationally feasible, current practices (e.g., DDPM~\citep{ho2020denoising} and SGM~\citep{song2021scorebased}) optimize the diffusion models by matching the backward process $p_{\theta}(\mathbf{x}_{0:T})$ with the forward process $q(\mathbf{x}_{0:T})$. This theorem indicates that this optimization scheme will fail for some complex data distribution $q(\mathbf{x}_0)$. 
    
    \textbf{Why diffusion models already perform well in practice.} The above theorem may bring unease---how can this be true when diffusion models are considered highly-realistic data generators? The key lies in the number of denoising steps. The more steps are used, the more the backward probability, Eq. (\ref{eq:backward proc}), is centered around a single mode, hence the more the simple Gaussian assumption holds~\citep{sohl2015deep}. As a result, we will see in Sec.~\ref{sec:inference speed} that our own method, which makes no Gaussian posterior assumption, improves quality especially for few backward iterations.
	
	\begin{mdframed}[leftmargin=0pt, rightmargin=0pt, innerleftmargin=10pt, innerrightmargin=10pt, skipbelow=0pt]
		\textbf{\emph{Takeaway}}: Standard diffusion models (e.g. DDPM) with simple Gaussian denoising poorly approximate some multimodal distributions (e.g. Gaussian mixture). This is problematic, as these distributions are very common in practice.
	\end{mdframed}

\section{Method: Soft Mixture Denoising}

	Our theoretical studies showed how current diffusion models have limited expressiveness to approximate multimodal data distributions. To solve this problem, we propose \textbf{soft mixture denoising} (SMD), a tractable relaxation of a Gaussian mixture model for modelling the denoising posterior.
	
\subsection{Main Theory}

	Our theoretical analysis highlight an expressive bottleneck of current diffusion models due to its Gaussian denoising assumption. Based on Proposition~\ref{lemma:posterior form}, an obvious way to address this problem is to directly model the backward probability $p_{\theta}(\mathbf{x}_{t - 1} \mid \mathbf{x}_t)$ as a Gaussian mixture. For example, we could model:
	\begin{equation}
		\label{eq:mixture backward}
		p_{\theta}^{\mathrm{mixture}}(\mathbf{x}_{t - 1} \mid \mathbf{x}_t) = \sum_{k=1}^K z_{\theta_k}(\mathbf{x}_t, t) \mathcal{N}(\mathbf{x}_{t - 1}; \bm{\mu}_{\theta_k}(\mathbf{x}_t, t), \bm{\Sigma}_{\theta_k}(\mathbf{x}_t, t)),
	\end{equation}
	where $\theta = \bigcup_{k=1}^K \theta_k$, the number of Gaussian components $K$ is a hyperparameter, and where weight $z^k_t(\cdot)$, mean $\bm{\mu}_{\theta_k}^k(\cdot)$, and covariance $\bm{\Sigma}_{\theta_k}^k(\cdot)$ are learnable and determine each of the mixture components. While the mixture model might be complex enough for backward denoising, it is not practical for two reasons: 1) it is often intractable to determine the number of components $K$ from observed data; 2) mixture models are notoriously hard to optimize. Actually, \cite{jin2016local} proved that a Gaussian mixture model might be optimized into an arbitrarily bad local optimum.
	
	\parag{Soft mixture denoising.} To efficiently improve the expressiveness of diffusion models, we introduce \textit{soft mixture denoising} (SMD) $p_{\widebar{\theta}}^{\mathrm{SMD}}(\mathbf{x}_{t-1} \mid \mathbf{x}_t)$, a soft version of the mixture model $p_\theta^{\mathrm{mixture}}(\cdot)$, which avoids specifying the number of mixture components $K$ and permits effective optimization. Specifically, we define a continuous latent variable $\mathbf{z}_t$, as an alternative to mixture weight $z^k_t$, that represents the potential mixture structure of posterior distribution $q(\mathbf{x}_{t-1} \mid \mathbf{x}_t)$. Under this scheme, we model the learnable backward probability as
	\begin{equation}
		\label{eq:backward redef}
			p_{\widebar{\theta}}^{\mathrm{SMD}} (\cdot) = \int p_{\widebar{\theta}}^{\mathrm{SMD}}(\mathbf{x}_{t-1}, \mathbf{z}_t \mid \mathbf{x}_t) d\mathbf{z}_t = \int p_{\widebar{\theta}}^{\mathrm{SMD}}(\mathbf{z}_t \mid \mathbf{x}_t) p_{\widebar{\theta}}^{\mathrm{SMD}}(\mathbf{x}_{t-1} \mid \mathbf{x}_t, \mathbf{z}_t) d\mathbf{z}_t,
	\end{equation}
	where $\widebar{\theta}$ denotes the set of all learnable parameters. We model $p_{\widebar{\theta}}(\mathbf{x}_{t-1} \mid \mathbf{x}_t, \mathbf{z}_t)$ as a learnable multivariate Gaussian and expect that different values of the latent variable $\mathbf{z}_t$ will correspond to differently parameterized Gaussians: 
	\begin{equation}
		p_{\widebar{\theta}}^{\mathrm{SMD}}(\mathbf{x}_{t-1} \mid \mathbf{x}_t, \mathbf{z}_t) = \mathcal{N}\big(\mathbf{x}_{t-1}; \bm{\mu}_{\theta\bigcup f_{\phi}(\mathbf{z}_t, t)}\big(\mathbf{x}_t, t\big), \bm{\Sigma}_{\theta\bigcup f_{\phi}( \mathbf{z}_t, t)}\big(\mathbf{x}_t, t\big)\big),
	\end{equation}
        where $\theta \subset \widebar{\theta}$ is a set of vanilla learnable parameters and $f_{\phi}(\mathbf{z}_t, t)$ is another collection of parameters computed from a neural network $f_{\phi}$ with learnable parameters $\phi \subset \widebar{\theta}$. Both $\theta$ and $f_{\phi}(\mathbf{z}_t, t)$ constitute the parameter set of mean and covariance functions $\bm{\mu}_{\bullet}, \bm{\Sigma}_{\bullet}$ for computations, but only $\theta$ and $\phi$ will be optimized. This type of design is similar to the hypernetwork~\citep{ha2017hypernetworks,krueger2018bayesian}. For implementation, we follow Eq.~(\ref{eq:backward proc}) to constrain the covariance matrix $\bm{\Sigma}_{\bullet}$ to the form $\sigma_t \mathbf{I}$ and parameterize mean $\bm{\mu}_{\bullet}(\mathbf{x}_t, t)$ similar to Eq.~(\ref{eq:mean def}):
	\begin{equation}
		\label{eq:new mean parameterization}
		\bm{\mu}_{\theta \bigcup  f_{\phi}(\mathbf{z}_t, t)}(\mathbf{x}_t, t) = \frac{1}{\sqrt{\alpha_t}} \Big(\mathbf{x}_t - \frac{\beta_t}{\sqrt{1 - \widebar{\alpha}_t}}\bm{\epsilon}_{\theta \bigcup f_{\phi}(\mathbf{z}_t, t)}\big(\mathbf{x}_t, t\big) \Big),
	\end{equation}
	where $\bm{\epsilon}_{\bullet}$ is a neural network. For image data, we build it as a U-Net~\citep{ronneberger2015u} (i.e., $\theta$) with several extra layers that are computed from $f_{\phi}(\mathbf{z}_t, t)$.

	For the mixture component $p_{\widebar{\theta}}(\mathbf{z}_t \mid \mathbf{x}_t)$, we parameterize it with a neural network such that it can be an arbitrarily complex distribution and adds great flexibility into the backward probability $p_{\widebar{\theta}}^{\mathrm{SMD}}(\mathbf{x}_{t-1} \mid \mathbf{x}_t)$.  For implementation, we adopt a mapping $g_{\xi}:(\bm{\eta},\mathbf{x}_t,t) \mapsto \mathbf{z}_t,\xi \subset \widebar{\theta}$ with $\bm{\eta}\overset{\mathrm{i.i.d.}}{\sim} \mathcal{N}(\mathbf{0}, \mathbf{I})$, which converts a standard Gaussian into a non-Gaussian distribution.
	
	\parag{Theoretical guarantee.} We prove that SMD $p_{\widebar{\theta}}^{\mathrm{SMD}}(\mathbf{x}_{t-1} \mid \mathbf{x}_t)$ improves the expressiveness of diffusion models---resolving the limitations highlighted in Theorems~\ref{prop:upper bound}~and~\ref{theorem:inconsistent backward proc}.
	
	\begin{theorem}[Expressive Soft Mixture Denoising]
		\label{theorem:zero error measure}
		
		For the diffusion process defined in Eq.~(\ref{eq:forward def}), suppose soft mixture model $p_{\widebar{\theta}}^{\mathrm{SMD}}(\mathbf{x}_{t-1} \mid \mathbf{x}_t)$ is applied for backward denoising and data distribution $q(\mathbf{x}_0)$ is a Gaussian mixture, then both $\mathcal{M}_t = 0, \forall t \in [1, T]$ and  $\mathcal{E} = 0$ hold.
	
		\begin{proof}
			The proof to this theorem is fully provided in Appendix~\ref{appendix:zero error measure}.
		\end{proof}
		
	\end{theorem}

		\begin{figure}[t]
		\begin{minipage}[t]{0.495\textwidth}
			\begin{algorithm}[H]
				\caption{Training} \label{alg:training}
				\small
				\begin{algorithmic}[1]
					\Repeat
					\State $\mathbf{x}_0 \sim q(\mathbf{x}_0)$
					\State $t \sim \mathcal{U}\{1, T\}$, $\bm{\epsilon} \sim \mathcal{N}(\mathbf{0}, \mathbf{I})$
					\State $\mathbf{x}_t = \sqrt{\widebar{\alpha}_t}\mathbf{x}_0 + \sqrt{1 - \widebar{\alpha}_t} \bm{\epsilon}$
                    \State \colorbox{blue!10}{$\bm{\eta} \sim \mathcal{N}(\mathbf{0}, \mathbf{I})$}
					\State \colorbox{blue!10}{Latent variable sampling: $\mathbf{z}_t = g_{\xi}(\bm{\eta}, \mathbf{x}_t, t)$}
					\State  \colorbox{blue!10}{Param. for computation: $\widehat{\theta} = \theta\bigcup f_{\phi}(\mathbf{z}_t, t)$}
					\State \colorbox{blue!10}{Param. to optimize: $\widebar{\theta} = \theta \bigcup \phi \bigcup  \xi$}
					\State Update $\widebar{\theta}$ w.r.t. $\nabla_{\widebar{\theta}}\|\bm{\epsilon} - \highlight{\bm{\epsilon}_{\widehat{\theta}}}(\mathbf{x}_t, t)\|^2$
					\Until{converged}
				\end{algorithmic}
			\end{algorithm}
		\end{minipage}
		\hfill
		\begin{minipage}[t]{0.495\textwidth}
			\begin{algorithm}[H]
				\caption{Sampling} \label{alg:sampling}
				\small
				\begin{algorithmic}[1]
					\vspace{.04in}
					\State $\mathbf{x}_T \sim p(\mathbf{x}_T) = \mathcal{N}(\mathbf{0}, \mathbf{I})$
					\For{$t=T, \dotsc, 1$}
					\State $\bm{\epsilon} \sim \mathcal{N}(\mathbf{0}, \mathbf{I})$ if $t > 1$, else $\bm{\epsilon} = 0$
                    \State \colorbox{blue!10}{$\bm{\eta} \sim \mathcal{N}(\mathbf{0}, \mathbf{I})$}
					\State \colorbox{blue!10}{Latent variable sampling: $\mathbf{z}_t = g_{\xi}(\bm{\eta}, \mathbf{x}_t, t)$}
					\State \colorbox{blue!10}{Param. for computation: $\widehat{\theta} = \theta \bigcup f_{\phi}(\mathbf{z}_t, t)$}
					\State $\mathbf{x}_{t-1} = \frac{1}{\sqrt{\alpha}_t}\Big(\mathbf{x}_t - \frac{1 - \alpha_t}{\sqrt{1 - \widebar{\alpha}_t}} \highlight{\bm{\epsilon}_{\widehat{\theta}}}(\mathbf{x}_t, t)\Big) + \sigma_t \bm{\epsilon}$
					\EndFor
					\State \textbf{return} $\mathbf{x}_0$
					\vspace{.04in}
				\end{algorithmic}
			\end{algorithm}
		\end{minipage}
		% \vspace{-1em}
	\end{figure}
    
    \begin{remark}
        The Gaussian mixture is a universal approximator for continuous probability distributions~\citep{dalal1983approximating}. Therefore, this theorem implies that our proposed SMD permits the diffusion models to well approximate arbitrarily complex data distributions.
    \end{remark}

    \begin{mdframed}[leftmargin=0pt, rightmargin=0pt, innerleftmargin=10pt, innerrightmargin=10pt, skipbelow=0pt]
		\textbf{\emph{Takeaway}}: Soft mixture denoising (SMD) parameterizes the backward probability as a continuously relaxed Gaussian mixture, which potentially permits the diffusion models to well approximate any continuous data distribution.
	\end{mdframed}
    
\subsection{Efficient Optimization and Sampling}

	While Theorem~\ref{theorem:zero error measure} shows that SMDs are highly expressive, it assumes the neural networks are globally optimized. Plus, the latent variable in SMD introduces more complexity to the computation and analysis of diffusion models. To fully exploit the potential of SMD, we thus need efficient optimization and sampling algorithms.

	\parag{Loss function.} The negative log-likelihood for a diffusion model with the backward probability $p^{\mathrm{SMD}}_{\widebar{\theta}}(\mathbf{x}_{t-1} \mid \mathbf{x}_t)$ of a latent variable model is formally defined as
	\begin{equation}
		\mathbb{E}_{q}[-\ln p_{\widebar{\theta}}^{\mathrm{SMD}}(\mathbf{x}_0)] = \mathbb{E}_{\mathbf{x}_0 \sim q(\mathbf{x}_0)} \Big[-\ln \Big(  \int_{\mathbf{x}_{1:T}} p(\mathbf{x}_T) \prod_{t=T}^{1} p^{\mathrm{SMD}}_{\widebar{\theta}}(\mathbf{x}_{t-1} \mid \mathbf{x}_t)  d\mathbf{x}_{1:T} \Big) \Big].
	\end{equation}
	Like vanilla diffusion models, this log-likelihood term is also computationally infeasible. In the following, we derive its upper bound for optimization.
	
	\begin{proposition}[Upper Bound of Negative Log-likelihood]
		\label{prop:new loss}
		
		Suppose the diffusion process is defined as Eq.~(\ref{eq:forward def}) and the soft mixture model $p_{\widebar{\theta}}^{\mathrm{SMD}}(\mathbf{x}_{t-1} \mid \mathbf{x}_t)$ is applied for backward denoising, then an upper bound of the expected negative log-likelihood $\mathbb{E}_{q}[-\ln p_{\widebar{\theta}}^{\mathrm{SMD}}(\mathbf{x}_0)]$ is
		\begin{equation}
			\label{eq:new approx loss}
			\mathcal{L}^{\mathrm{SMD}} = C + \sum_{t=1}^{T}  \mathbb{E}_{\bm{\eta}, \bm{\epsilon}, \mathbf{x}_0} \Big[ \Gamma_t \big\| \bm{\epsilon} - \bm{\epsilon}_{\theta \bigcup  f_{\phi}(g_\xi(\cdot), t)} \big(\sqrt{\widebar{\alpha}_t} \mathbf{x}_0 + \sqrt{1 - \widebar{\alpha}_t} \bm{\epsilon}, t\big)  \big\|^2 \Big],
		\end{equation}
		where $g_\xi(\cdot) = g_\xi(\bm{\eta}, \sqrt{\widebar{\alpha}_t} \mathbf{x}_0 + \sqrt{1 - \widebar{\alpha}_t} \bm{\epsilon}, t)$, $C$ is a constant that does not involve any learnable parameter $\widebar{\theta} = \theta\bigcup \phi \bigcup \xi$, $\mathbf{x}_0 \sim q(\mathbf{x}_0)$, $\bm{\eta}, \bm{\epsilon}$ are two independent variables drawn from standard Gaussians, and $\Gamma_t = \beta_t^2 / (2\sigma_t\alpha_t(1 - \widebar{\alpha}_t))$.
		\begin{proof}
			The detailed derivation to get the upper bound $\mathcal{L}^{\mathrm{SMD}}$ is in Appendix~\ref{appendix:new upper bound}.
		\end{proof}
		
\end{proposition}

	Compared with the loss function of vanilla diffusion models, Eq.~(\ref{eq:simplified loss}), our upper bound mainly differs in the hypernetwork $f_{\phi}$ to parameterize the denoising function $\bm{\epsilon}_{\bullet}$ and an expectation operation $\mathbb{E}_{\bm{\eta}}$. The former is computed by neural networks and the latter is approximated by Monte Carlo sampling, which both add minor computational costs.
	
	\parag{Training and Inference.} The SMD training and sampling procedures are respectively shown in Algorithms~\ref{alg:training} and \ref{alg:sampling}, with blue highlighting differences with vanilla diffusion. For the training procedure, we follow common practices of ~\citep{ho2020denoising,NEURIPS2021_49ad23d1}, and (1) apply Monte Carlo sampling to handle iterated expectations $\mathbb{E}_{\bm{\eta}, \bm{\epsilon}, \mathbf{x}_0}$ in Eq.~(\ref{eq:new approx loss}), and (2) reweigh loss term $\|\bm{\epsilon} - \bm{\epsilon}_{\bullet}(\mathbf{x}_t, t)\|^2$ by ignoring coefficient $\Gamma_t$. 
 One can also sample more noises (e.g., $\bm{\eta}$) in one training step to trade run-time efficiency for approximation accuracy.

\section{Experiments}

Let us verify how SMD improves the quality and speed of existing diffusion models. First, we use a toy example to visualise that existing diffusion models struggle to learn multivariate Gaussians, whereas SMD does not. Subsequently, we show how SMD significantly improves the FID score across different types of diffusion models (e.g., DDPM, ADM~\citep{NEURIPS2021_49ad23d1}, LDM) and datasets. Then, we demonstrate how SMD significantly improves performance at low number of inference steps. This enables reducing the number of inference steps, thereby speeding up generation and reducing computational costs. Lastly, we show how quality can be improved even further by sampling more than one $\bm{\eta}$ for loss estimation at training time, which further improves the performance but causes an extra time cost.

\subsection{Visualising the Expressive Bottleneck}
\begin{figure}
  \centering
    \includegraphics[width=1.0\linewidth]{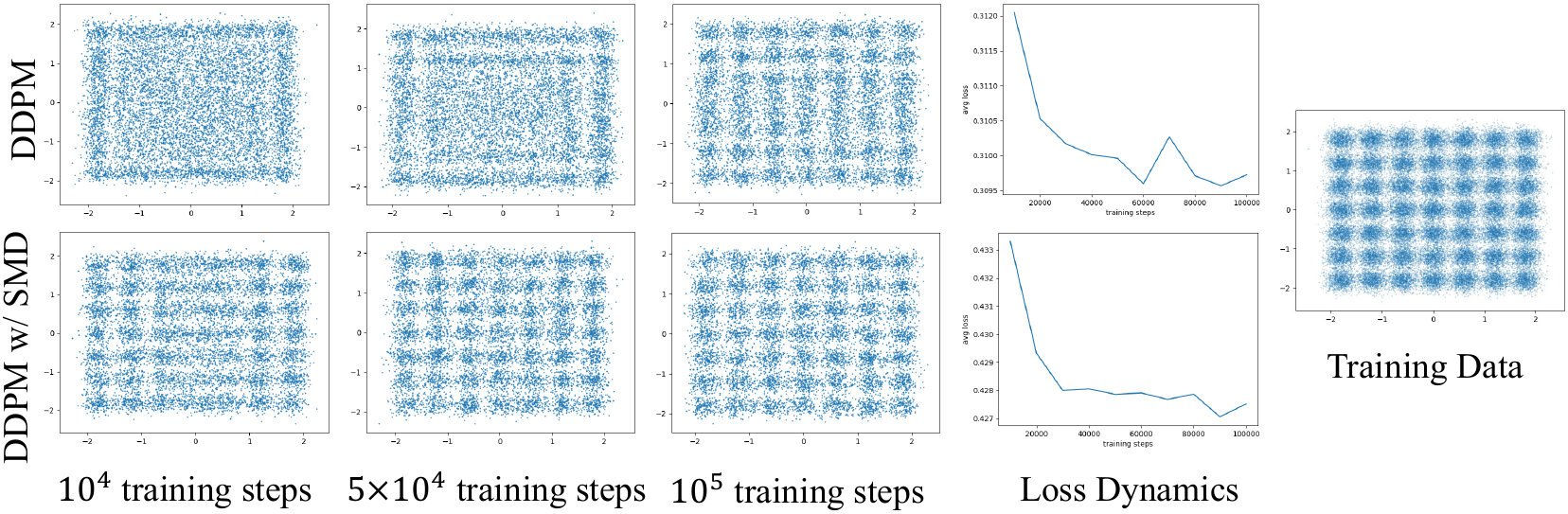}
  \caption{\textbf{Visualising the expressive bottleneck of standard diffusion models.} Experimental results on synthetic dataset with $7\times 7$ Gaussians (right), for DDPM with $T=1000$. Even though DDPM has converged, we observe that the modes are not easily distinguishable. On the other hand, SMD converges much faster and results in distinguishable modes.}
  \label{fig:toy_experiment}
\end{figure}

From Proposition \ref{lemma:posterior form} and Theorems \ref{theorem:inconsistent backward proc}, \ref{prop:upper bound}  it follows that vanilla diffusion models would struggle with learning a Gaussian Mixture model, whereas Theorem \ref{theorem:zero error measure} proves SMD does not. Let us visualise this difference using a simple toy experiment. In Figure \ref{fig:toy_experiment} we plot the learnt distribution of DDPM over the training process, with and without SMD. We observe that DDPM with SMD converges much faster, and provides a more accurate distribution at time of convergence.

		\begin{table}[bt]
		\centering
		\caption{\textbf{SMD consistently improves generation quality.} FID score of different models across common image datasets and resolutions. We use $T = 1000$ for all models.}
		\begin{tabular}{ccc|cc}
			
			\hline
			
			Dataset / Model & DDPM & DDPM w/ SMD & ADM & ADM w/ SMD \\
			
			\hline
			
			CIFAR-10 ($32 \times 32$) &  $3.78$ & $\mathbf{3.13}$ &  $2.98$ & $\mathbf{2.55}$ \\
			
			LSUN-Conference ($64 \times 64$) & $4.15$ & $\mathbf{3.52}$ & $3.85$ & $\mathbf{3.29}$ \\
			
			LSUN-Church ($64 \times 64$) & $3.65$ & $\mathbf{3.17}$ & $3.41$ & $\mathbf{2.98}$ \\
			
			CelebA-HQ ($128 \times 128$) & $6.78$ & $\mathbf{6.35}$ & $6.45$ & $\mathbf{6.02}$ \\
			
			\hline
			
		\end{tabular}
		\label{tab:real_world_mid}
  \vspace{-3mm}
	\end{table}

\subsection{SMD Improves Image Quality}

	We select three of the most common diffusion models and four image datasets to show how our proposed SMD quantitatively improves diffusion models. Baselines include DDPM~\cite{ho2020denoising}, ADM~\citep{NEURIPS2021_49ad23d1}, and Latent Diffusion Model (LDM)~\citep{pinaya2022brain}. Datasets include CIFAR-10~\citep{krizhevsky2009learning}, LSUN-Conference, LSUN-Church~\citep{yu2015lsun}, and CelebA-HQ~\citep{liu2015faceattributes}. For all models, we set the backward iterations $T$ as $1000$ and generate $10000$ images for computing FID scores.

 	\begin{wraptable}{r}{0.6\textwidth}
		\centering
  \vspace{-3mm}
		\caption{\textbf{SMD improves LDM generation quality.} FID score of latent diffusion with and without SMD on high-resolution image datasets ($T = 1000$).}
    \begin{tabular}{ccc} \hline
    Dataset / Model                & LDM    & LDM w/ SMD      \\ \hline
    LSUN-Church ($256 \times 256$) & $5.86$ & $\mathbf{5.21}$ \\
    CelebA-HQ ($256 \times 256$)   & $6.13$ & $\mathbf{5.48}$ \\ \hline
    \end{tabular}
    		\label{tab:real_world_high}
    	\end{wraptable}
	In Table~\ref{tab:real_world_mid}, we show how the proposed SMD significantly improves both DDPM and ADM on all datasets, for a range of resolutions. For example, SDM outperforms DDPM by $15.14\%$ on LSUN-Church and ADM by $16.86\%$. Second, in Table~\ref{tab:real_world_high} we include results for high-resolution image datasets, see Fig.~\ref{fig:demo} for example images ($T=100$). Here we employed LDM as baseline to reduce memory footprint, where we use a pretrained and frozen VAE. We observe that SMD improves FID scores significantly. These results strongly indicate how SMD is effective in improving the performance for different baseline diffusion models.

\subsection{SMD Improves Inference Speed}
\label{sec:inference speed}
	\begin{wrapfigure}{r}{0.5\textwidth}
	\centering
 \vspace{-3mm}
			\includegraphics[width=0.5\textwidth]{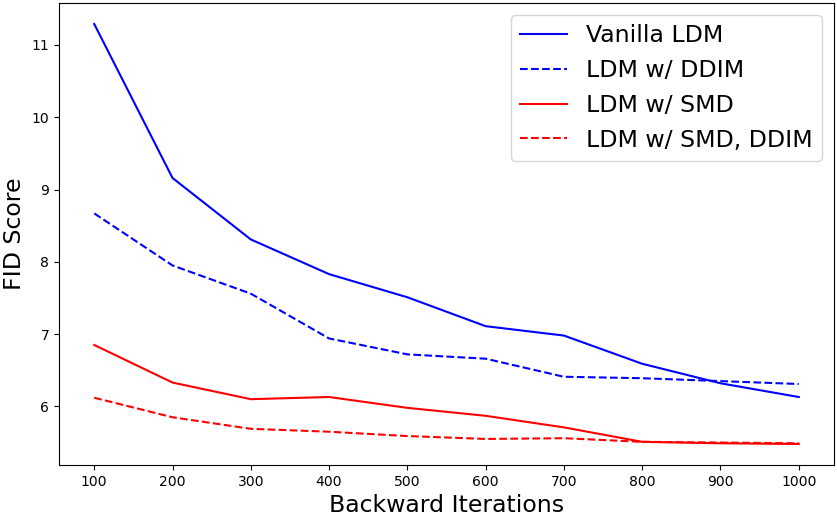}
			\caption{\textbf{SMD reduces the number of sampling steps.} Latent DDIM and DDPM for different iterations on CelebA-HQ ($256 \times 256$).}
               \vspace{-3mm}
			\label{fig:few iters}
	\end{wrapfigure}%
	Intuitively, for few denoising iterations the distribution $q(\mathbf{x}_{t-1} \mid \mathbf{x}_t)$ is more of a mixture, which leads to the backward probability $p_{\theta}(\mathbf{x}_{t-1} \mid \mathbf{x}_t)$---a simple Gaussian---being a worse approximation. Based on Theorems~\ref{theorem:inconsistent backward proc}~and~\ref{theorem:zero error measure}, we anticipate that our models will be more robust to this effect than vanilla diffusion models. 
	
	The solid blue and red curves in Fig.~\ref{fig:few iters} respectively show how the F1 scores of vanilla LDM and LDM w/ SMD change with respect to increasing backward iterations. We can see that our proposed SMD improves the LDM much more at fewer backward iterations (e.g., $T = 200$). We also include LDM with DDIM~\citep{song2021denoising}, a popular fast sampler. We see that the advantage of SDM is consistent across samplers.
		
\subsection{Sampling Multiple $\eta$: a Cost-Quality Trade-off}
\begin{wrapfigure}{r}{0.5\textwidth}
			    \vspace{-3mm}
               \centering\includegraphics[width=0.5\textwidth]{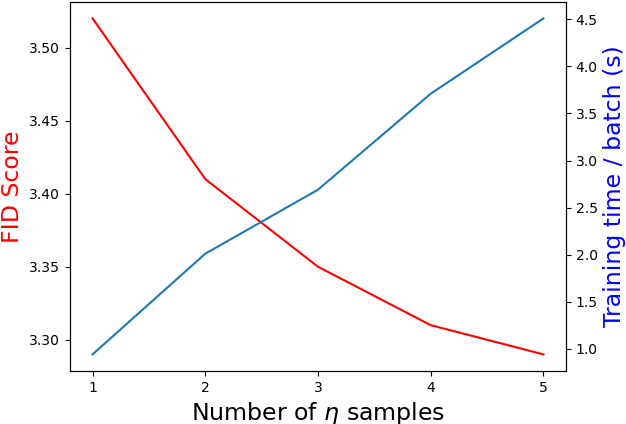}
			\caption{\textbf{SMD quality is further improved by sampling multiple $\eta$,} see Alg.~\ref{alg:training} on LSUN-Conference ($64 \times 64$) for DDPM w/ SMD.}
            \vspace{-3mm}
			\label{fig:eta}
	\end{wrapfigure}
	In Algorithm~\ref{alg:training}, we only sample one $\bm{\eta}$ at a time for maintaining high computational efficiency. We can sample multiple $\eta$ to estimate the loss better. Figure \ref{fig:eta} shows how the training time of one training step and FID score of DDPM with SMD changes as a function of the number of $\eta$ samples. While the time cost linearly goes up with the increasing sampling times, FID monotonically decreases (6.5\% for 5 samples).

\section{Future Work}
We have proven that there exists an expressive bottleneck in popular diffusion models. Since multimodal distributions are so common, this limitation does matter across domains (e.g., tabular, images, text). Our proposed SMD, as a general method for expressive backward denoising, solves this problem. Regardless of network architectures, SMD can be extended to other tasks, including text-to-image translation and speech synthesis. Because SMD provides better quality for fewer steps, we also hope it will become a standard part of diffusion libraries, speeding up both training and inference.

\clearpage
\bibliography{iclr2024_conference}
\bibliographystyle{iclr2024_conference}

\appendix

\appendix

\section{Proof of Proposition \ref{lemma:posterior form}}
\label{appendix:proof to posterior}

By repeatedly applying basic operations (e.g., chain rule) of probability theory to conditional distribution of backward variable $q(\mathbf{x}_{t-1} \mid \mathbf{x}_t)$, we have
\begin{equation}
	\begin{aligned}
		q(\mathbf{x}_{t-1} \mid \mathbf{x}_{t}) & = \frac{q(\mathbf{x}_t, \mathbf{x}_{t-1})}{q(\mathbf{x}_t)} = \frac{q(\mathbf{x}_t \mid \mathbf{x}_{t-1}) q(\mathbf{x}_{t-1})}{q(\mathbf{x}_{t})} = \frac{q(\mathbf{x}_t \mid \mathbf{x}_{t-1})}{q(\mathbf{x}_{t})} \int_{\mathbf{x}_0} q(\mathbf{x}_{t-1}, \mathbf{x}_0) d\mathbf{x}_0 \\
		& = \frac{1}{q(\mathbf{x}_{t})} q(\mathbf{x}_t \mid \mathbf{x}_{t-1}) \int_{\mathbf{x}_0} q(\mathbf{x}_{t-1} \mid \mathbf{x}_0) q(\mathbf{x}_0) d\mathbf{x}_0
	\end{aligned}.
\end{equation}
Based on Eq.~(\ref{eq:forward def}) and $q(\mathbf{x}_t \mid \mathbf{x}_0) = \mathcal{N}(\mathbf{x}_t; \sqrt{\widebar{\alpha}_t} \mathbf{x}_0, (1 - \widebar{\alpha}_t)\mathbf{I})$, from \citep{ho2020denoising}, posterior probability $q(\mathbf{x}_{t-1} \mid \mathbf{x}_t)$ can be expressed as
\begin{equation}
	\label{eq:raw reverse}
	q(\mathbf{x}_{t-1} \mid \mathbf{x}_{t}) = \frac{\mathcal{N}(\mathbf{x}_t; \sqrt{1 - \beta_t} \mathbf{x}_{t-1}, \beta_t \mathbf{I})}{q(\mathbf{x}_t)} \int_{\mathbf{x}_0} \mathcal{N}(\mathbf{x}_{t-1}; \sqrt{\widebar{\alpha}_{t-1}} \mathbf{x}_0, (1 - \widebar{\alpha}_{t-1})\mathbf{I}) q(\mathbf{x}_0) d\mathbf{x}_0.
\end{equation}
Note that for a multivariate Gaussian, the following holds:
\begin{equation}
	\label{eq:scaled gaussian}
	\begin{aligned}
		\mathcal{N}(\mathbf{x}; \lambda \bm{\mu}, \bm{\Sigma}) & = (2\pi)^{-\frac{D}{2}}|\bm{\Sigma}|^{-\frac{1}{2}} \exp \Big(-\frac{1}{2}(\mathbf{x} - \lambda \bm{\mu})^T\bm{\Sigma}^{-1}(\mathbf{x} - \lambda \bm{\mu})\Big) \\
		& = \frac{1}{\lambda^{D}} (2\pi)^{-\frac{D}{2}}  \Big| \frac{\bm{\Sigma}}{\lambda^2} \Big|^{-\frac{1}{2}}  \exp \Big(-\frac{1}{2}\big(\bm{\mu} - \frac{\mathbf{x}}{\lambda} \big)^T\big( \frac{\bm{\Sigma}}{\lambda^2} \big)^{-1}\big(\bm{\mu} - \frac{\mathbf{x}}{\lambda} \big) \Big) \\
		& = (1 / \lambda)^D  \mathcal{N} (\bm{\mu}; \mathbf{x} / \lambda, \bm{\Sigma} / \lambda^2 ) 
	\end{aligned},
\end{equation}
where $\lambda \in \mathbb{R}^+$, $\bm{\mu}$ denotes a vector with dimension $D$, and $\bm{\Sigma}$ is a positive semi-definite matrix. Fromt that, and $\beta_t = 1 - \alpha_t$, the following identities follow:
\begin{equation}
	\left\{\begin{aligned}
		\mathcal{N}(\mathbf{x}_t; \sqrt{1 - \beta_t} \mathbf{x}_{t-1}, \beta_t \mathbf{I}) & = \alpha_t^{-\frac{D}{2}} \mathcal{N} \Big(\mathbf{x}_{t-1}; \frac{\mathbf{x}_t}{\sqrt{\alpha_t}},  \frac{1 - \alpha_t}{\alpha_t} \mathbf{I} \Big) \\
		\mathcal{N}(\mathbf{x}_{t-1}; \sqrt{\widebar{\alpha}_{t-1}} \mathbf{x}_0, (1 - \widebar{\alpha}_{t-1})\mathbf{I})  & = (\widebar{\alpha}_{t-1})^{-\frac{D}{2}} \mathcal{N} \Big( \mathbf{x}_0; \frac{\mathbf{x}_{t-1}}{\sqrt{\widebar{\alpha}_{t-1}} },  \frac{1 - \widebar{\alpha}_{t-1}}{\widebar{\alpha}_{t-1}}\mathbf{I} \Big)
	\end{aligned}\right..
\end{equation}
Therefore, we can refomulate Eq.~(\ref{eq:raw reverse}) as
\begin{equation}
	\label{eq:new reverse}
	q(\cdot) = \frac{(\alpha_t\widebar{\alpha}_{t-1})^{-\frac{D}{2}} }{q(\mathbf{x}_t)} \mathcal{N} \Big(\mathbf{x}_{t-1}; \frac{\mathbf{x}_t}{\sqrt{\alpha_t}},  \frac{1 - \alpha_t}{\alpha_t} \mathbf{I} \Big) \int_{\mathbf{x}_0} \mathcal{N} \Big( \mathbf{x}_0; \frac{\mathbf{x}_{t-1}}{\sqrt{\widebar{\alpha}_{t-1}} },  \frac{1 - \widebar{\alpha}_{t-1}}{\widebar{\alpha}_{t-1}}\mathbf{I} \Big) q(\mathbf{x}_0) d\mathbf{x}_0.
\end{equation}
Now, we let $q(\mathbf{x}_0)$ be a mixture of Gaussians $q(\mathbf{x}_0) = \sum_{k=1}^K w_k \mathcal{N}(\mathbf{x}_0; \bm{\mu}_k, \bm{\Sigma}_k)$, where $K$ is the number of Gaussian components, $w_k \in [0, 1]$, $\sum_k w_k = 1$, and vector $\bm{\mu}_k$ and matrix $\bm{\Sigma}_k$ respectively denote the mean and covariance of component $k$.

For the the mixture of Gaussians distribution $q(\mathbf{x}_0)$ and by exchanging the operation order of summation $\sum_{k=1}^K$ and integral $\int_{\mathbf{x}_0}$, we have
\begin{equation}
	\label{eq:mixture reverse}
	\begin{aligned}
		q(\mathbf{x}_{t-1} \mid \mathbf{x}_{t}) & = \sum_{k=1}^K \Big[  \frac{w_k(\alpha_t\widebar{\alpha}_{t-1})^{-\frac{D}{2}}}{q(\mathbf{x}_t)} \mathcal{N} \Big(\mathbf{x}_{t-1}; \frac{\mathbf{x}_t}{\sqrt{\alpha_t}},  \frac{1 - \alpha_t}{\alpha_t} \mathbf{I} \Big)   \\ 
		& * \int_{\mathbf{x}_0} \mathcal{N} \Big( \mathbf{x}_0; \frac{\mathbf{x}_{t-1}}{\sqrt{\widebar{\alpha}_{t-1}} },  \frac{1 - \widebar{\alpha}_{t-1}}{\widebar{\alpha}_{t-1}}\mathbf{I} \Big) \mathcal{N} \Big( \mathbf{x}_0; \bm{\mu}_k, \bm{\Sigma}_k \Big) d\mathbf{x}_0 \Big].
	\end{aligned}
\end{equation}
A nice property of Gaussian distributions is that the product of two multivariate Gaussians also follows a Gaussian distribution~\citep{ahrendt2005multivariate}. Formally, we have
\begin{equation}
	\label{eq:gaussian product}
	\begin{aligned}
		& \mathcal{N}(\mathbf{x}; \bm{\mu}_1, \bm{\Sigma}_1) \mathcal{N}(\mathbf{x}; \bm{\mu}_2, \bm{\Sigma}_2) = \mathcal{N}(\bm{\mu}_2; \bm{\mu}_1, \bm{\Sigma}_1 + \bm{\Sigma}_2) \\
		& * \mathcal{N}(\mathbf{x}; (\bm{\Sigma}_1^{-1} + \bm{\Sigma}_2^{-1})^{-1} (\bm{\Sigma}_1^{-1} \bm{\mu}_1 + \mathbf{\Sigma}_2^{-1} \bm{\mu}_2 ), (\bm{\Sigma}_1^{-1} + \bm{\Sigma}_2^{-1})^{-1} )
	\end{aligned},
\end{equation}
where $\bm{\mu}_1, \bm{\mu}_2$ are vectors of the same dimension and $\bm{\Sigma}_1, \bm{\Sigma}_2$ are positive-definite matrices. Therefore, the integral part $\int_{\mathbf{x}_0}$ in Eq.~(\ref{eq:mixture reverse}) can be computed as
\begin{equation}
	\begin{aligned}
		& \int_{\mathbf{x}_0} \mathcal{N} \Big( \mathbf{x}_0; \frac{\mathbf{x}_{t-1}}{\sqrt{\widebar{\alpha}_{t-1}} },  \frac{1 - \widebar{\alpha}_{t-1}}{\widebar{\alpha}_{t-1}}\mathbf{I} \Big) \mathcal{N} \Big( \mathbf{x}_0; \bm{\mu}_k, \bm{\Sigma}_k \Big) d\mathbf{x}_0 \\ 
		& = \mathcal{N} \Big( \bm{\mu}_k; \frac{\mathbf{x}_{t-1}}{\sqrt{\widebar{\alpha}_{t-1}} }, \frac{1 - \widebar{\alpha}_{t-1}}{\widebar{\alpha}_{t-1}}\mathbf{I} + \bm{\Sigma}_k  \Big) * \int_{\mathbf{x}_0} \mathcal{N}(\mathbf{x}_0; \cdot, \cdot) d\mathbf{x}_0 \\
		& = (\widebar{\alpha}_{t-1})^{-\frac{D}{2}} \mathcal{N} (\mathbf{x}_{t-1}; \sqrt{\widebar{\alpha}_{t-1}} \bm{\mu}_k, (1 - \widebar{\alpha}_{t-1})\mathbf{I} + \widebar{\alpha}_{t-1} \bm{\Sigma}_k  ) * 1
	\end{aligned},
\end{equation}
where the last equation is derived by Eq.~(\ref{eq:scaled gaussian}). With this result, we have
\begin{equation}
	q(\mathbf{x}_{t-1} \mid \mathbf{x}_{t}) = \sum_{k=1}^K \Big[ \frac{w_k\alpha_t^{-\frac{D}{2}}}{q(\mathbf{x}_t)} \mathcal{N} \Big(\cdot\Big) \mathcal{N} \Big(\mathbf{x}_{t-1}; \sqrt{\widebar{\alpha}_{t-1}} \bm{\mu}_k, (1 - \widebar{\alpha}_{t-1})\mathbf{I} + \widebar{\alpha}_{t-1} \bm{\Sigma}_k \Big)  \Big],
\end{equation}
By applying Eq.~(\ref{eq:gaussian product}) and Eq.~(\ref{eq:scaled gaussian}), and $\widebar{\alpha}_{t-1}\alpha_t = \widebar{\alpha}_{t}$, the product of two Gaussian distributions in the above equality can be reformulated as
\begin{equation}
	\begin{aligned}
		& \mathcal{N} \Big(\mathbf{x}_{t-1}; \frac{\mathbf{x}_t}{\sqrt{\alpha_t}},  \frac{1 - \alpha_t}{\alpha_t} \mathbf{I} \Big) * \mathcal{N} \Big(\mathbf{x}_{t-1}; \sqrt{\widebar{\alpha}_{t-1}} \bm{\mu}_k, (1 - \widebar{\alpha}_{t-1})\mathbf{I} + \widebar{\alpha}_{t-1} \bm{\Sigma}_k \Big) \\
		& = \alpha_{t}^{\frac{D}{2}} \mathcal{N} \Big( \mathbf{x}_t; \sqrt{ \widebar{\alpha}_{t}} \bm{\mu}_k, (1 -  \widebar{\alpha}_{t}) \mathbf{I} + \widebar{\alpha}_{t} \bm{\Sigma}_k  \Big) \\
		& * \mathcal{N}  \Big(  \mathbf{x}_{t-1}; (\mathbf{I} + \bm{\Lambda}_k^{-1})^{-1} \frac{\mathbf{x}_t}{\sqrt{\alpha_t}} + (\mathbf{I} + \bm{\Lambda}_k)^{-1} \sqrt{ \widebar{\alpha}_{t-1}} \bm{\mu}_k, \frac{1 - \alpha_t}{\alpha_t} (\mathbf{I} + \bm{\Lambda}_k^{-1})^{-1}  \Big) 
	\end{aligned},
\end{equation}
where matrix $\bm{\Lambda}_k = (\alpha_t - \widebar{\alpha}_t) / (1 - \alpha_t) \mathbf{I} +  \widebar{\alpha}_t / (1 - \alpha_t) \bm{\Sigma}_k$. With this result, we have
\begin{equation}
    \label{eq:posterior form}
	\left\{\begin{aligned}
		q(\mathbf{x}_{t-1} \mid \mathbf{x}_{t}) & = \sum_{k=1}^K w_k' \mathcal{N}  ( \mathbf{x}_{t-1}; \bm{\mu}_k',  \bm{\Sigma}_k') \\
		w_k' & = \frac{w_k}{q(\mathbf{x}_t)} \mathcal{N} ( \mathbf{x}_t; \sqrt{ \widebar{\alpha}_{t}} \bm{\mu}_k, (1 -  \widebar{\alpha}_{t}) \mathbf{I} + \widebar{\alpha}_{t} \bm{\Sigma}_k) \\
		\bm{\mu}_k' & = (\mathbf{I} + \bm{\Lambda}_k^{-1})^{-1} \frac{\mathbf{x}_t}{\sqrt{\alpha_t}} + (\mathbf{I} + \bm{\Lambda}_k)^{-1} \sqrt{ \widebar{\alpha}_{t-1}} \bm{\mu}_k \\
		\bm{\Sigma}_k' & =  \frac{1 - \alpha_t}{\alpha_t} (\mathbf{I} + \bm{\Lambda}_k^{-1})^{-1}
	\end{aligned}\right.,
\end{equation}
where $\sum_{k=1}^K w'_k = 1$. To conclude, from this equality it follows that posterior probability $p(\mathbf{x}_{t-1} \mid \mathbf{x}_t)$ is also a mixture of Gaussians. Therefore, our proposition holds.

\section{Proof of Theorem \ref{prop:upper bound}}
\label{appendix:upper bound}

	Let us rewrite metric $\mathcal{M}_t$ as
	\begin{equation}
		\label{eq:decomposition of error metric}
		\begin{aligned}
			\mathcal{M}_t & = \inf_{\theta \in \Theta} \Big( \int_{\mathbf{x}_t} q(\mathbf{x}_t) \big( \int_{\mathbf{x}_{t-1}} q(\mathbf{x}_{t-1} \mid \mathbf{x}_t) \ln \frac{q(\mathbf{x}_{t-1} \mid \mathbf{x}_t)}{p_{\theta}(\mathbf{x}_{t-1} \mid \mathbf{x}_t)} d\mathbf{x}_{t-1} \big) d\mathbf{x}_t \Big) \\
			& = \inf_{\theta \in \Theta} \Big( \int_{\mathbf{x}_t} q(\mathbf{x}_t) \big( -\mathcal{H}[q(\mathbf{x}_{t-1} \mid \mathbf{x}_t)] + \mathcal{D}_{\mathrm{CE}}[q(\mathbf{x}_{t-1} \mid \mathbf{x}_t), p_{\theta}(\mathbf{x}_{t-1} \mid \mathbf{x}_t)] \big) d\mathbf{x}_t \Big)
		\end{aligned},
	\end{equation}
	where $\mathcal{H}[\cdot]$ is information entropy~\citep{shannon2001mathematical}:
	\begin{equation}
		\mathcal{H}[q(\mathbf{x}_{t-1} \mid \mathbf{x}_t)] = -\int_{\mathbf{x}_{t-1}} q(\mathbf{x}_{t-1} \mid \mathbf{x}_t) \ln q(\mathbf{x}_{t-1} \mid \mathbf{x}_t) d\mathbf{x}_{t-1},
	\end{equation}
	and $\mathcal{D}_{\mathrm{CE}}[\cdot]$ denotes the cross-entropy~\citep{de2005tutorial}:
	\begin{equation}
		\mathcal{D}_{\mathrm{CE}}[q(\mathbf{x}_{t-1} \mid \mathbf{x}_t), p_{\theta}(\mathbf{x}_{t-1} \mid \mathbf{x}_t)] =  - \int_{\mathbf{x}_{t-1}} q(\mathbf{x}_{t-1} \mid \mathbf{x}_t) \ln p_{\theta}(\mathbf{x}_{t-1} \mid \mathbf{x}_t) d\mathbf{x}_{t-1}.
	\end{equation}
	Note that the entropy term $\mathcal{H}[\cdot]$ does not involve parameter $\theta$ and can be regarded as a normalization term for adjusting the minimum of $\mathcal{D}_{\mathrm{KL}}[\cdot]$ to $0$.

Our goal is to analyze error metric $\mathcal{M}_t$ defined in Eq.~(\ref{eq:def of error metric}). Regarding its decomposition derived in Eq.~(\ref{eq:decomposition of error metric}), we first focus on cross-entropy $\mathcal{D}_{\mathrm{CE}}[q(\mathbf{x}_{t-1} \mid \mathbf{x}_t), p_{\theta}(\mathbf{x}_{t-1} \mid \mathbf{x}_t)]$. Suppose $q(\mathbf{x}_0)$ follows a Gaussian mixture, then $q(\mathbf{x}_{t-1} \mid \mathbf{x}_t)$ is also such a distribution as formulated in Eq.~(\ref{eq:lemma gaussian mixture}). Therefore, we can expand the above cross entropy $\mathcal{D}_{\mathrm{CE}}$ as
\begin{equation}
	\begin{aligned}
		\mathcal{D}_{\mathrm{CE}}[\cdot] & = - \int_{\mathbf{x}_{t-1}} q(\mathbf{x}_{t-1} \mid \mathbf{x}_t) \ln p_{\theta}(\mathbf{x}_{t-1} \mid \mathbf{x}_t) d\mathbf{x}_{t-1} \\
		& = - \int_{\mathbf{x}_{t-1}} \Big( \sum_{k=1}^K w_k' \mathcal{N}  ( \mathbf{x}_{t-1}; \bm{\mu}_k',  \bm{\Sigma}_k') \Big) \ln p_{\theta}(\mathbf{x}_{t-1} \mid \mathbf{x}_t) d\mathbf{x}_{t-1} \\
		& = \sum_{k=1}^K w_k' \mathcal{D}_{\mathrm{CE}}[\mathcal{N}  ( \mathbf{x}_{t-1}; \bm{\mu}_k',  \bm{\Sigma}_k'), p_{\theta}(\mathbf{x}_{t-1} \mid \mathbf{x}_t)] \\
		& = \sum_{k=1}^K w_k' \mathcal{D}_{\mathrm{KL}}[\mathcal{N}  ( \mathbf{x}_{t-1}; \bm{\mu}_k',  \bm{\Sigma}_k'), p_{\theta}(\mathbf{x}_{t-1} \mid \mathbf{x}_t)] + \sum_{k=1}^K w_k' \mathcal{H}[\mathcal{N}  ( \mathbf{x}_{t-1}; \bm{\mu}_k',  \bm{\Sigma}_k')]
	\end{aligned}.
\end{equation}
Suppose we set $\bm{\Sigma}_k = \delta_k \mathbf{I}, \delta_k > 0$, then we have
\begin{equation}
	\left\{\begin{aligned}
		\bm{\mu}'_k & = \Big(\frac{1 + (\delta_k - 1)\widebar{\alpha}_{t-1}}{1 + (\delta_k - 1)\widebar{\alpha}_{t}}\Big) \sqrt{\alpha_t} \mathbf{x}_t + \frac{(1 - \alpha_t)\sqrt{\widebar{\alpha}_{t-1}}}{1 + (\delta_k - 1)\widebar{\alpha}_{t}}\bm{\mu}_k \\
		\bm{\Sigma}'_k & = \Big(\frac{1 + (\delta_k - 1)\widebar{\alpha}_{t-1}}{1 + (\delta_k - 1)\widebar{\alpha}_{t}}\Big) (1 - \alpha_t) \mathbf{I}
	\end{aligned}\right..
\end{equation}
With this equation, we can simplify entropy sum $\sum_{k=1}^K w_k' \mathcal{H}[\cdot]$ as
\begin{equation}
	\sum_{k=1}^K w_k' \mathcal{H}[\mathcal{N}  ( \mathbf{x}_{t-1}; \bm{\mu}_k',  \bm{\Sigma}_k')  = \sum_{k=1}^K \frac{w_k'}{2}  \ln | 2\pi\mathrm{e}  \bm{\Sigma}'_k | = \frac{D}{2} \ln (2\pi \mathrm{e}) + \sum_{k=1}^K \frac{w_k'}{2} \ln | \bm{\Sigma}'_k |. \\ 
\end{equation}
Term $\mathcal{D}_{\mathrm{KL}}[\cdot]$ is in fact the KL divergence between two multivariate Gaussians, $\mathcal{N}  ( \mathbf{x}_{t-1}; \bm{\mu}_k',  \bm{\Sigma}'_k )$ and $\mathcal{N}(\mathbf{x}_{t-1}; \bm{\mu}_{\theta}(\mathbf{x}_t, t), \sigma_t\mathbf{I})$, which has an analytic form~\citep{zhang2021properties}:
\begin{equation}
	\begin{aligned}
		& \mathcal{D}_{\mathrm{KL}}[\cdot]  = \frac{1}{2}\Big( \ln \frac{|\sigma_t\mathbf{I}|}{|\bm{\Sigma}'_k |} - D + \frac{1}{\sigma_t}\|\bm{\mu}_k' - \bm{\mu}_{\theta}(\mathbf{x}_t, t)\|^2 + \mathrm{Tr}\{(\sigma_t\mathbf{I})^{-1} \bm{\Sigma}'_k  \} \Big) \\
		& =  \frac{1}{2} \Big( D \ln \sigma_t - \ln |\bm{\Sigma}'_k| - D \Big) + \frac{1}{2\sigma_t}\|\bm{\mu}_k' - \bm{\mu}_{\theta}(\mathbf{x}_t, t)\|^2 + \frac{1 - \alpha_t}{2\sigma_t} \frac{1 + (\delta_k - 1)\widebar{\alpha}_{t-1}}{1 + (\delta_k - 1)\widebar{\alpha}_{t}} D
	\end{aligned}.
\end{equation}
With the above two equalities and the fact that $\widebar{\alpha}_{t-1} > \widebar{\alpha}_{t}$ because $\alpha_t < 1$, we reduce term $\mathcal{D}_{\mathrm{CE}}[q(\mathbf{x}_{t-1} \mid \mathbf{x}_t), p_{\theta}(\mathbf{x}_{t-1} \mid \mathbf{x}_t)]$ as
\begin{equation}
	\label{eq:final def of ce}
	\mathcal{D}_{\mathrm{CE}}[\cdot] > \frac{1}{2\sigma_t} \sum_{k=1}^K w_k' \|\bm{\mu}_k' - \bm{\mu}_{\theta}(\mathbf{x}_t, t)\|^2 + \frac{D}{2} \ln(2\pi\sigma_t) + \frac{1 - \alpha_t}{2\sigma_t} D.
\end{equation}
Since entropy $\mathcal{H}[q(\mathbf{x}_{t-1} \mid \mathbf{x}_t)]$ does not involve model parameter $\theta$, the variation of error metric $\mathcal{M}_t$ is from cross-entropy $\mathcal{D}_{\mathrm{CE}}[\cdot]$, more specifically, sum $\sum_{k=1}^K$. Let's focus on how this term contributes to error metric $\mathcal{M}_t$ as formulated in Eq.~(\ref{eq:def of error metric}):
\begin{equation}
	\mathcal{I}_{\mathrm{CE}} =  \int_{\mathbf{x}_t} q(\mathbf{x}) \sum_{k=1}^K w_k' \|\bm{\mu}_k' - \bm{\mu}_{\theta}(\mathbf{x}_t, t)\|^2 d\mathbf{x}_t = \sum_{k=1}^K \Big( \int_{\mathbf{x}_t} w_k' q(\mathbf{x}) \|\bm{\mu}_k' - \bm{\mu}_{\theta}(\mathbf{x}_t, t)\|^2 d\mathbf{x}_t \Big).
\end{equation}
Considering that Eq.~(\ref{eq:lemma gaussian mixture}) and $\bm{\Sigma}_k$ has been set as $\delta_k \mathbf{I}$, we have
\begin{equation}
	\begin{aligned}
		\mathcal{I}_{\mathrm{CE}} & = \sum_{k=1}^K \Big( \int_{\mathbf{x}_t} w_k \mathcal{N} \Big( \mathbf{x}_t; \sqrt{ \widebar{\alpha}_{t}} \bm{\mu}_k, (1 + (\delta_k - 1)\widebar{\alpha}_{t}) \mathbf{I} \Big) \Big\|\bm{\mu}_k' - \bm{\mu}_{\theta}(\mathbf{x}_t, t) \Big\|^2 d\mathbf{x}_t \Big) \\
		& =  \int_{\mathbf{x}_t} \mathcal{N} ( \cdot ) \Big( \sum_{k=1}^K w_k \Big\| \Big(\frac{(1 - \alpha_t)\sqrt{\widebar{\alpha}_{t-1}}}{1 + (\delta_k - 1)\widebar{\alpha}_{t}} \Big) \bm{\mu}_k - \Big(\bm{\mu}_{\theta}(\mathbf{x}_t, t) - \Big(\cdot\Big) \sqrt{\alpha_t} \mathbf{x}_t  \Big) \Big\|^2 \Big) d\mathbf{x}_t \\
	\end{aligned}.
\end{equation}
Sum $\sum_{k=1}^K w_k \| \cdot \|^2$ is essentially a problem called weighted least squares~\citep{rousseeuw2005robust} for model $\bm{\mu}_{\theta}(\mathbf{x}_t, t) - (\cdot)\sqrt{\alpha_t}\mathbf{x}_t$, which achieves a minimum error when the model is $\sum_{k=1}^K w_k (\cdot) \bm{\mu}_k$. For convenience, we suppose $\sum_{k=1}^K w_k \bm{\mu}_k / (1 + (\delta_k - 1)\widebar{\alpha}_{t}) = \mathbf{0}$ and we have
\begin{equation}
	\label{eq:cross-entropy contribution}
	\mathcal{I}_{\mathrm{CE}} \ge \Big(\int_{\mathbf{x}_t} \mathcal{N} ( \cdot ) d\mathbf{x}_t \Big) \Big( \sum_{k=1}^K w_k \Big\|\Big(\cdot\Big) \bm{\mu}_k \Big\|^2 \Big) = (1 - \alpha_t)^2 \widebar{\alpha}_{t-1} \sum_{k=1}^K w_k \Big\|\frac{\bm{\mu}_k}{1 + (\delta_k - 1)\widebar{\alpha}_{t}} \Big\|^2.
\end{equation}
Term $\mathcal{H}[q(\mathbf{x}_{t-1} \mid \mathbf{x}_t)]$ is in fact the differential entropy of a Gaussian mixture. Considering our previous setup and its upper bound provided by \citep{huber2008entropy}, we have
\begin{equation}
	\begin{aligned}
		\mathcal{H}[\cdot] & \le \sum_{k=1}^K w_k' \Big(-\ln w_k' + \frac{1}{2}\ln \Big((2\pi\mathrm{e})^D\Big| \frac{1 + (\delta_k - 1)\widebar{\alpha}_{t-1}}{1 + (\delta_k - 1)\widebar{\alpha}_{t}} (1 - \alpha_t) \mathbf{I}\Big|\Big)\Big) \\
		& < \frac{D}{2} \ln\Big(\frac{2\pi \mathrm{e} }{\alpha_t} (1 - \alpha_t)\Big) -\sum_{k=1}^K w_k' \ln w_k'  \le \frac{D}{2} \ln\Big(2\pi \mathrm{e} \Big(\frac{1}{\alpha_t} - 1\Big)\Big) + \ln K
	\end{aligned},
\end{equation}
where the second ineqaulity holds since $(1 + x) / (1 + xy) < 1 / y, \forall x \in \mathbb{R}^+, y \in (0, 1)$ and the last inequality is obtained by regarding term $-\sum_{k=1}^K$ as the entropy of discrete variables $[w_1', w_2', \cdots, w_K']$. Therefore, its contribution to error metric $\mathcal{M}_t$ is
\begin{equation}
	\mathcal{I}_{\mathrm{Ent}} =  \int_{\mathbf{x}_t} q(\mathbf{x}_t) (-\mathcal{H}[q(\mathbf{x}_{t-1} \mid \mathbf{x}_t)]) d\mathbf{x}_t \ge -\frac{D}{2} \ln\Big(\frac{2\pi \mathrm{e} }{\alpha_t} (1 - \alpha_t)\Big) - \ln K.
\end{equation}
Combining this inequality with Eq.~(\ref{eq:final def of ce}) and Eq.~(\ref{eq:cross-entropy contribution}), we have
\begin{equation}
	\mathcal{M}_t > \frac{ (1 - \alpha_t)^2 \widebar{\alpha}_{t-1}}{2\sigma_t} \sum_{k=1}^K w_k \Big\|\frac{\bm{\mu}_k}{1 + (\delta_k - 1)\widebar{\alpha}_{t}} \Big\|^2 - \ln K + \frac{D}{2}\Big( \ln \frac{\sigma_t \alpha_t}{1 - \alpha_t} + \frac{1 - \alpha_t}{\sigma_t} - 1 \Big).
\end{equation}
with constraint $\sum_{k=1}^K w_k \bm{\mu}_k / (1 + (\delta_k - 1)\widebar{\alpha}_{t}) = \mathbf{0}$. Since $w_k > 0, 1 \le k \le K$, there exists a group of non-zero vectors $[\bm{\mu}_1, \bm{\mu}_2, \cdots, \bm{\mu}_K]$ satisfying this linear equation, corresponds to a Gaussian mixture $p(\mathbf{x}_0)$. With this result, we can always find another group of solution $[\lambda \bm{\mu}_1, \lambda \bm{\mu}_2, \cdots, \lambda \bm{\mu}_K]$ for $\lambda \in \mathbb{R}$, which corresponds to a new mixture of Gaussians. By increasing the value of $\lambda$, the first term of this inequality can be arbitrarily and uniformly large in terms of iteration $t$.

\section{Proof of Theorem~\ref{theorem:inconsistent backward proc}}
\label{proof:inconsistent backward proc}

	Due to the first-order markov property of the forward and backward processes and the fact $q(\mathbf{x}_T)= p_{\theta}(\mathbf{x}_T) = \mathcal{N}(\mathbf{0}, \mathbf{I}), T \rightarrow \infty$, we first have
	\begin{equation}
		\begin{aligned}
		& \mathcal{D}_{\mathrm{KL}}[\cdot] = \mathbb{E}_{\mathbf{x}_{0:T} \sim q(\mathbf{x}_{0:T})} \Big[ \ln  \frac{q(\mathbf{x}_{0:T})}{p_{\theta}(\mathbf{x}_{0:T})}  \Big]  = \mathbb{E}_q \Big[ \ln  \frac{q(\mathbf{x}_T) \prod_{t=T}^{1} q(\mathbf{x}_{t-1} \mid \mathbf{x}_t)}{p_{\theta}(\mathbf{x}_T) \prod_{t=T}^{1} p_{\theta} (\mathbf{x}_{t-1} \mid \mathbf{x}_t)}  \Big] \\
		& = \mathbb{E}_q \Big[ \sum_{t = 1}^T \ln \frac{q(\mathbf{x}_{t-1} \mid \mathbf{x}_t)}{p_{\theta} (\mathbf{x}_{t-1} \mid \mathbf{x}_t)} \Big] = \sum_{t = 1}^T E_{\mathbf{x}_t} \Big[ \mathcal{D}_{\mathrm{KL}}[q(\mathbf{x}_{t-1} \mid \mathbf{x}_t), p_{\theta} (\mathbf{x}_{t-1} \mid \mathbf{x}_t)] \Big]
		\end{aligned},
	\end{equation}
	where the last equality holds because of the following derivation:
	\begin{equation}
		\begin{aligned}
			& \mathbb{E}_q \Big[ \ln \frac{q(\mathbf{x}_{t-1} \mid \mathbf{x}_t)}{p_{\theta} (\mathbf{x}_{t-1} \mid \mathbf{x}_t)} \Big] = \int_{\mathbf{x}_{0:T}} q(\mathbf{x}_{0:T}) \ln \frac{q(\mathbf{x}_{t-1} \mid \mathbf{x}_t)}{p_{\theta} (\mathbf{x}_{t-1} \mid \mathbf{x}_t)} d \mathbf{x}_{0:T} \\
			& = \int_{\mathbf{x}_{t-1}} q(\mathbf{x}_t) \Big( \int_{\mathbf{x}_{t}} q(\mathbf{x}_{t-1} \mid \mathbf{x}_t) \ln \frac{q(\mathbf{x}_{t-1} \mid \mathbf{x}_t)}{p_{\theta} (\mathbf{x}_{t-1} \mid \mathbf{x}_t)}  d\mathbf{x}_{t-1} \Big) d\mathbf{x}_t \\
			& = E_{\mathbf{x}_t \sim q(\mathbf{x}_t)} \Big[ \mathcal{D}_{\mathrm{KL}}[q(\mathbf{x}_{t-1} \mid \mathbf{x}_t), p_{\theta} (\mathbf{x}_{t-1} \mid \mathbf{x}_t)] \Big].
		\end{aligned}
	\end{equation}
	Based on Theorem~\ref{prop:upper bound}, then we can infer that there is a continuous data distribution $q(\mathbf{x}_0)$ such that the inequality $\mathcal{M}_t > (N + 1) / T$ holds for $t \in [1, T]$. For this distribution, we have
	\begin{equation}
		 \mathcal{D}_{\mathrm{KL}}[\cdot] \ge  \sum_{t = 1}^T \inf \Big( E_{\mathbf{x}_t} \Big[ \mathcal{D}_{\mathrm{KL}}[q(\mathbf{x}_{t-1} \mid \mathbf{x}_t), p_{\theta} (\mathbf{x}_{t-1} \mid \mathbf{x}_t)] \Big] \Big) = \sum_{t = 1}^T M_t  > N + 1.
	\end{equation}
	Finally, we get $\mathcal{E} = \inf(\mathcal{D}_{\mathrm{KL}}[\cdot]) \ge N + 1 > N$ for the data distribution $q(\mathbf{x}_0)$.

\section{Proof of Theorem~\ref{theorem:zero error measure}}
\label{appendix:zero error measure}

	We split the proof into two parts: one for $\mathcal{M}_t, t \in [1, T]$ and the other for $\mathcal{E}$.

	\paragraph{Zero local denoising errors.} For convenience, we denote integral $\int_{\mathbf{x}_t} q(\mathbf{x}_t) \mathcal{D}_{\mathrm{KL}}[\cdot] d\mathbf{x}_t$ in the definition of error measure $\mathcal{M}_t$ as $\mathcal{M}_t (\widebar{\theta})$. Immediately, we have $\mathcal{M}_t = \inf_{\widebar{\theta} \in \widebar{\Theta}} \mathcal{M}_t(\widebar{\theta})$. With this equality, it suffices to prove two assertions: $\mathcal{M}_t(\widebar{\theta}) \ge 0, \forall \widebar{\theta} \in \Theta$ and $\exists \widebar{\theta} \in \widebar{\Theta}: \mathcal{M}_t(\widebar{\theta}) = 0.$
	
	The first assertion is trivially true since KL divergence $\mathcal{D}_{\mathrm{KL}}$ is always non-negative. For the second assertion, we introduce two lemmas: 1) The assertion is true for the mixture model $p_{\theta}^{\mathrm{mixture}}(\mathbf{x}_{t - 1} \mid \mathbf{x}_t)$; 2) Any mixture model can be represented by its soft version $p_{\widebar{\theta}}^{\mathrm{SMD}} (\mathbf{x}_{t-1}  \mid \mathbf{x}_t)$. If we can prove the two lemma, it is sufficient to say that the second assertion also holds for SMD.
	
	We prove the first lemma by construction. According to Proposition~\ref{lemma:posterior form}, the inverse forward probability $q(\mathbf{x}_{t-1} \mid \mathbf{x}_t)$ is also a Gaussian mixture as formulated in Eq.~(\ref{eq:posterior form}). By selecting a proper number $K$, the mixture model $p_{\theta}^{\mathrm{mixture}}(\mathbf{x}_{t - 1} \mid \mathbf{x}_t)$ defined in Eq.~(\ref{eq:mixture backward}) will be of the same distribution family as its reference $q(\mathbf{x}_{t-1} \mid \mathbf{x}_t)$, which only differ in the configuration of different mixture components. Based on Eq.~(\ref{eq:posterior form}), we can specifically set parameter $\theta = \bigcup_{1 \le k \le K} \theta_k$ as
	\begin{equation}
		\left\{\begin{aligned}
			z_{\theta_k}(\mathbf{x}_t, t) & ~\propto~ w_k \mathcal{N} ( \mathbf{x}_t; \sqrt{ \widebar{\alpha}_{t}} \bm{\mu}_k, (1 -  \widebar{\alpha}_{t}) \mathbf{I} + \widebar{\alpha}_{t} \bm{\Sigma}_k) \\
			\bm{\mu}_{\theta_k}(\mathbf{x}_t, t) & = (\mathbf{I} + \bm{\Lambda}_k^{-1})^{-1} \frac{\mathbf{x}_t}{\sqrt{\alpha_t}} + (\mathbf{I} + \bm{\Lambda}_k)^{-1} \sqrt{ \widebar{\alpha}_{t-1}} \bm{\mu}_k \\
			\bm{\Sigma}_{\theta_k}(\mathbf{x}_t, t) & =  \frac{1 - \alpha_t}{\alpha_t} (\mathbf{I} + \bm{\Lambda}_k^{-1})^{-1} \\
			\bm{\Lambda}_k & = \frac{\alpha_t - \widebar{\alpha}_t}{1 - \alpha_t} \mathbf{I} + \frac{\widebar{\alpha}_t}{1 - \alpha_t} \bm{\Sigma}_k
		\end{aligned}\right.,
	\end{equation}
	such that the backward probability $p_{\theta}^{\mathrm{mixture}}(\mathbf{x}_{t - 1} \mid \mathbf{x}_t)$ is the same as its reference $q(\mathbf{x}_{t-1} \mid \mathbf{x}_t)$ and thus $\mathcal{D}_{\mathrm{KL}}[q(\mathbf{x}_{t-1} \mid \mathbf{x}_t), p_{\theta}^{\mathrm{mixture}}(\mathbf{x}_{t - 1} \mid \mathbf{x}_t)]$ by definition is $0$. In this sense, we also have $\mathcal{M}_t(\theta) = 0$, which exactly proves the first lemma.
	
	We also prove the second lemma by construction. Given any mixture model $p_{\theta}^{\mathrm{mixture}}(\mathbf{x}_{t - 1} \mid \mathbf{x}_t)$ as defined in Eq.~(\ref{eq:mixture backward}), we divide the space $\mathbb{R}^L$ (where $L$ is the vector dimension of variable $\mathbf{z}_t$) into $K$ disjoint subsets $\{\mathcal{Z}_{t,1}, \mathcal{Z}_{t,2}, \cdots, \mathcal{Z}_{t, K}\}$ such that:
	\begin{equation}
		\int_{\mathbf{z}_t \in \mathcal{Z}_{t, k}} p_{\widebar{\theta}}^{\mathrm{SMD}}(\mathbf{z}_t \mid \mathbf{x}_t)  d\mathbf{z}_t = z_{\theta_k}(\mathbf{x}_t, t), \ \ \ \ \theta_k = f_{\phi}( \mathbf{z}_t, t), \forall \mathbf{z}_t \in \mathcal{Z}_{t, k},
	\end{equation}
	where $k \in \{1,..., K\}$. The first equality can be true for any continuous density $p_{\widebar{\theta}}^{\mathrm{SMD}}$ and the second one can be implemented by a simple step function. By setting $\theta = \emptyset$, we have
	\begin{equation}
		\begin{aligned}
			p_{\widebar{\theta}}^{\mathrm{SMD}} & (\mathbf{x}_{t-1}  \mid \mathbf{x}_t) = \int_{\mathbf{z}_t} p_{\widebar{\theta}}^{\mathrm{SMD}}(\mathbf{z}_t \mid \mathbf{x}_t) \mathcal{N}(\mathbf{x}_{t-1}; \bm{\mu}_{\theta, f_{\phi}(\mathbf{z}_t, t)}(\mathbf{x}_t, t), \bm{\Sigma}_{\theta, f_{\phi}( \mathbf{z}_t, t)}(\mathbf{x}_t, t)) d\mathbf{z}_t \\ & = \sum_{k=1}^{K} \Big( \int_{\mathbf{z}_t \in \mathcal{Z}_{t, k}} p_{\widebar{\theta}}^{\mathrm{SMD}}(\mathbf{z}_t \mid \mathbf{x}_t) \mathcal{N}\big(\mathbf{x}_{t-1}; \bm{\mu}_{f_{\phi}(\cdot)}(\mathbf{x}_t, t), \bm{\Sigma}_{f_{\phi}(\cdot)}(\mathbf{x}_t, t)\Big) d\mathbf{z}_t  \big) \\
			& =  \sum_{k=1}^{K} \Big( \mathcal{N}\big(\mathbf{x}_{t-1}; \bm{\mu}_{\theta_k}(\mathbf{x}_t, t), \bm{\Sigma}_{\theta_k}(\mathbf{x}_t, t)\big)  \int_{\mathbf{z}_t \in \mathcal{Z}_{t, k}} p_{\widebar{\theta}}^{\mathrm{SMD}}(\mathbf{z}_t \mid \mathbf{x}_t) d\mathbf{z}_t \Big) \\
			& = \sum_{k=1}^{K} \Big( \mathcal{N}(\mathbf{x}_{t - 1}; \bm{\mu}_{\theta_k}(\mathbf{x}_t, t), \bm{\Sigma}_{\theta_k}(\mathbf{x}_t, t)) z_{\theta_k}(\mathbf{x}_t, t) \Big) = p_{\theta}^{\mathrm{mixture}}(\mathbf{x}_{t - 1} \mid \mathbf{x}_t)
		\end{aligned},
	\end{equation}
	which actually proves the second lemma.
	
	\paragraph{Zero global denoising error.} We can see from above that there is always a properly parameterized backward probability $p_{\widebar{\theta}}^{\mathrm{SMD}}$ for any Gaussian mixture $q(\mathbf{x}_0)$ such that $q(\mathbf{x}_{t-1} \mid \mathbf{x}_t) = p_{\widebar{\theta}}^{\mathrm{SMD}}(\mathbf{x}_{t-1} \mid \mathbf{x}_t), \forall t \in [1, T]$. Considering $q(\mathbf{x}_T) = p_{\widebar{\theta}}^{\mathrm{SMD}}(\mathbf{x}_T)$, we have
	\begin{equation}
		p_{\widebar{\theta}}^{\mathrm{SMD}}(\mathbf{x}_{T-1}, \mathbf{x}_T)  = p_{\widebar{\theta}}^{SMD}(\mathbf{x}_{T}) p_{\widebar{\theta}}^{\mathrm{SMD}}( \mathbf{x}_{T-1} \mid \mathbf{x}_T) = q(\mathbf{x}_T) q(\cdot) = q(\mathbf{x}_{T-1}, \mathbf{x}_T).
	\end{equation}
	Immediately, we can get $q(\mathbf{x}_{T-1}) = p_{\widebar{\theta}}^{\mathrm{SMD}}(\mathbf{x}_{T-1})$ since
	\begin{equation}
		p_{\widebar{\theta}}^{\mathrm{SMD}}(\mathbf{x}_{T-1}) = \int_{\mathbf{x}_T} p_{\widebar{\theta}}^{\mathrm{SMD}}(\mathbf{x}_{T-1}, \mathbf{x}_T) \mathbf{x}_T = \int_{\mathbf{x}_T} q(\mathbf{x}_{T-1}, \mathbf{x}_T) \mathbf{x}_{T} = q(\mathbf{x}_{T-1}).
	\end{equation}
	With the above results, we can further prove that $p_{\widebar{\theta}}^{\mathrm{SMD}}(\mathbf{x}_{T-2}, \mathbf{x}_{T-1}, \mathbf{x}_T) = q(\mathbf{x}_{T-2}, \mathbf{x}_{T-1}, \mathbf{x}_T)$ and $p_{\widebar{\theta}}^{\mathrm{SMD}}(\mathbf{x}_{T-2}) = q(\mathbf{x}_{T-2})$. By iterating this process for the subscript $t$ from $T$ to $1$, we will finally have $p_{\widebar{\theta}}(\mathbf{x}_{0:T}) = q(\mathbf{x}_{0:T})$ such that $\mathcal{E} = 0$.
	
\section{Proof of Proposition \ref{prop:new loss}}
\label{appendix:new upper bound}

While we have introduced a new family of backward probability $p_{\widebar{\theta}}^{\mathrm{SMD}}(\mathbf{x}_{t-1} \mid \mathbf{x}_t)$ in Eq.~(\ref{eq:backward redef}), upper bound $\mathcal{L} = \sum_{t=0}^T \mathcal{L}_t$ defined in Eq.~(\ref{eq:jensen}) is still valid for deriving the loss function. To avoid confusion, we add a superscript $\mathrm{SMD}$ to new loss terms. An immediate conclusion is that $\mathcal{L}^{\mathrm{SMD}}_T = 0$ because $p(\mathbf{x}_t)$ by definition is a standard Gaussian and $q(\mathbf{x}_T \mid \mathbf{x}_0)$ also well approximates this distribution for large $T$. Therefore, the focus of this proof is on terms of KL divergence $\mathcal{L}_{t-1}^{\mathrm{SMD}}, 1 < t \le T$ and negative log-likelihood $\mathcal{L}_0^{\mathrm{SMD}}$.

Based on the fact that $q(\mathbf{x}_{t-1} \mid \mathbf{x}_t, \mathbf{x}_0)$ has a closed-form solution:
\begin{equation}
	q(\mathbf{x}_{t-1} \mid \mathbf{x}_t, \mathbf{x}_0) = \mathcal{N}(\mathbf{x}_{t-1}; \widetilde{\bm{\mu}}_t(\mathbf{x}_t, \mathbf{x}_0), \widetilde{\beta}_t \mathbf{I}),
\end{equation}
where mean $\widetilde{\bm{\mu}}_t(\mathbf{x}_t, \mathbf{x}_0)$ and variance $\widetilde{\beta}_t$ are respectively defined as
\begin{equation}
	\widetilde{\bm{\mu}}_t(\mathbf{x}_t, \mathbf{x}_0) = \frac{\sqrt{\widebar{\alpha}_{t-1}}\beta_t}{1 - \widebar{\alpha}_t}\mathbf{x}_0 + \frac{\sqrt{\alpha_t}(1 - \widebar{\alpha}_{t-1})}{1 - \widebar{\alpha}_t} \mathbf{x}_t, \ \ \ \widetilde{\beta}_t = \frac{1 - \widebar{\alpha}_{t-1}}{1 - \widebar{\alpha}_t} \beta_t,
\end{equation} 
we expand term $\mathcal{L}_{t-1}^{\mathrm{SMD}} = \mathbb{E}_q [D_{\mathrm{KL}} ( q(\mathbf{x}_{t-1} \mid \mathbf{x}_t, \mathbf{x}_0) \mid\mid p^{\mathrm{SMD}}_{\widebar{\theta}} (\mathbf{x}_{t-1} \mid \mathbf{x}_{t})  ) ]$ as
\begin{equation}
	\begin{aligned}
		\mathcal{L}_{t-1}^{\mathrm{SMD}}  & = \mathbb{E}_{\mathbf{x}_0, \mathbf{x}_t \sim q(\mathbf{x}_0) q(\mathbf{x}_t \mid \mathbf{x}_0)} \Big[ \int_{\mathbf{x}_{t-1}} q(\mathbf{x}_{t-1} \mid \mathbf{x}_t, \mathbf{x}_0) \ln \frac{q(\mathbf{x}_{t-1} \mid \mathbf{x}_t, \mathbf{x}_0)}{p_{\widebar{\theta}}^{\mathrm{SMD}} (\mathbf{x}_{t-1} \mid \mathbf{x}_t)} d\mathbf{x}_{t-1} \Big] \\
		& = \mathbb{E}_q \Big[ -\mathcal{H}[q(\mathbf{x}_{t-1} \mid \mathbf{x}_t, \mathbf{x}_0) \Big] + \mathcal{D}_{\mathrm{CE}} \Big[q(\mathbf{x}_{t-1} \mid \mathbf{x}_t, \mathbf{x}_0), p_{\widebar{\theta}}^{\mathrm{SMD}} (\mathbf{x}_{t-1} \mid \mathbf{x}_t) ] \Big]
	\end{aligned}.
\end{equation}
Considering our new definition of backward probability $p_{\widebar{\theta}}^{\mathrm{SMD}} (\mathbf{x}_{t-1} \mid \mathbf{x}_t)$ in Eq.~(\ref{eq:backward redef}) and applying Jensen's inequality, we can infer
\begin{equation}
	\begin{aligned}
		\mathcal{D}_{\mathrm{CE}}[\cdot] & = -\mathbb{E}_{\mathbf{x}_{t-1} \sim q(\mathbf{x}_{t-1} \mid \mathbf{x}_t, \mathbf{x}_0)} \Big[ \ln \int_{\mathbf{z}_t} p_{\widebar{\theta}}^{\mathrm{SMD}}(\mathbf{z}_t \mid \mathbf{x}_t) p_{\widebar{\theta}}^{\mathrm{SMD}}(\mathbf{x}_{t-1} \mid \mathbf{x}_t, \mathbf{z}_t) d\mathbf{z}_t \Big] \\
		& = -\mathbb{E}_{\mathbf{x}_{t-1} \sim q(\mathbf{x}_{t-1} \mid \mathbf{x}_t, \mathbf{x}_0)} \Big[\ln \mathbb{E}_{\mathbf{z}_t \sim p_{\widebar{\theta}}^{\mathrm{SMD}}(\mathbf{z}_t \mid \mathbf{x}_t)}[p_{\widebar{\theta}}^{\mathrm{SMD}}(\mathbf{x}_{t-1} \mid \mathbf{x}_t, \mathbf{z}_t) d\mathbf{z}_t] \Big] \\
		& \le -\mathbb{E}_{\mathbf{x}_{t-1} \sim q(\mathbf{x}_{t-1} \mid \mathbf{x}_t, \mathbf{x}_0)} \Big[ \mathbb{E}_{\mathbf{z}_t \sim p_{\widebar{\theta}}^{\mathrm{SMD}}(\mathbf{z}_t \mid \mathbf{x}_t)}[ \ln p_{\widebar{\theta}}^{\mathrm{SMD}}(\mathbf{x}_{t-1} \mid \mathbf{x}_t, \mathbf{z}_t) d\mathbf{z}_t] \Big] \\
		& =  \mathbb{E}_{\mathbf{z}_t \sim p_{\theta}^{\mathrm{SMD}}(\mathbf{z}_t \mid \mathbf{x}_t)} \Big[ -\int_{\mathbf{x}_{t-1}} q(\mathbf{x}_{t-1} \mid \mathbf{x}_t, \mathbf{x}_0) \ln p_{\widebar{\theta}}^{\mathrm{SMD}}(\mathbf{x}_{t-1} \mid \mathbf{x}_t, \mathbf{z}_t) d\mathbf{x}_{t-1} \Big] \\
		& =  \mathbb{E}_{\mathbf{z}_t \sim p_{\widebar{\theta}}^{\mathrm{SMD}}(\mathbf{z}_t \mid \mathbf{x}_t)} \Big[\mathcal{D}_{\mathrm{CE}} [q(\mathbf{x}_{t-1} \mid \mathbf{x}_t, \mathbf{x}_0), p_{\widebar{\theta}}^{\mathrm{SMD}}(\mathbf{x}_{t-1} \mid \mathbf{x}_t, \mathbf{z}_t)] \Big]
	\end{aligned}.
\end{equation}
Combining the above two equations, we have
\begin{equation}
	\begin{aligned}
		\mathcal{L}_{t-1}^{\mathrm{SMD}}  & \le \mathbb{E}_q \Big[ -\mathcal{H}[q(\mathbf{x}_{t-1} \mid \mathbf{x}_t, \mathbf{x}_0)] + \mathbb{E}_{\mathbf{z}_t} [\mathcal{D}_{\mathrm{CE}} [q(\mathbf{x}_{t-1} \mid \mathbf{x}_t, \mathbf{x}_0), p_{\widebar{\theta}}^{\mathrm{SMD}}(\mathbf{x}_{t-1} \mid \mathbf{x}_t, \mathbf{z}_t)] ] \Big] \\
		& = \mathbb{E}_{q, \mathbf{z}_t} \Big[ -\mathcal{H}[q(\mathbf{x}_{t-1} \mid \mathbf{x}_t, \mathbf{x}_0)] + \mathcal{D}_{\mathrm{CE}} [q(\mathbf{x}_{t-1} \mid \mathbf{x}_t, \mathbf{x}_0), p_{\widebar{\theta}}^{\mathrm{SMD}}(\mathbf{x}_{t-1} \mid \mathbf{x}_t, \mathbf{z}_t)] \Big] \\
		& = \mathbb{E}_{\mathbf{z}_t} \Big[\mathbb{E}_{\mathbf{x}_0, \mathbf{x}_t \sim q(\mathbf{x}_0) q(\mathbf{x}_t \mid \mathbf{x}_0)} [ \mathcal{D}_{\mathrm{KL}} [q(\mathbf{x}_{t-1} \mid \mathbf{x}_t, \mathbf{x}_0), p_{\widebar{\theta}}^{\mathrm{SMD}} (\mathbf{x}_{t-1} \mid \mathbf{x}_{t}, \mathbf{z}_t) ] ] \Big] 
	\end{aligned}.
\end{equation}
Considering $\mathbf{z}_t = g_{\varphi}(\bm{\eta}, \mathbf{x}_t, t)$ and applying the law of the unconscious statistician (LOTUS)~\citep{pmlr-v37-rezende15}, we can simplify the above inequality as
\begin{equation}
	\mathcal{L}_{t-1}^{\mathrm{SMD}}  \le \mathbb{E}_{\bm{\eta} \sim \mathcal{N}(\mathbf{0}, \mathbf{I})} \big[ \mathbb{E}_q [\mathcal{D}_{\mathrm{KL}} [ q(\mathbf{x}_{t-1} \mid \mathbf{x}_t, \mathbf{x}_0), p_{\widebar{\theta}}^{\mathrm{SMD}} (\mathbf{x}_{t-1} \mid \mathbf{x}_{t}, g_{\xi}(\bm{\eta}, \mathbf{x}_t, t) ] ] \big].
\end{equation}
The inner term of expectation $\mathbb{E}_{\bm{\eta} \sim \mathcal{N}(\mathbf{0}, \mathbf{I})}[\cdot]$ is essentially the same as the old definition of $\mathcal{L}^{\mathrm{SMD}} _t$ in Eq.~(\ref{eq:jensen}), except that term $p_{\widebar{\theta}}(\cdot)$ is additionally conditional on $\mathbf{z}_t$. Hence, we follow the procedure of DDPM~\cite{ho2020denoising} to reduce it. The result is given without proving:
\begin{equation}
	\label{eq:new loss part-1}
	\left\{\begin{aligned}
		\mathcal{L}_{t-1}^{\mathrm{SMD}} & \le C_t + \mathbb{E}_{\bm{\eta}, \bm{\epsilon}, \mathbf{x}_0} \Big[ \frac{\beta_t^2}{2\sigma_t\alpha_t(1 - \widebar{\alpha}_t)}  \| \bm{\epsilon} - \bm{\epsilon}_{\theta, f_{\phi}(\cdot)} (\sqrt{\widebar{\alpha}_t} \mathbf{x}_0 + \sqrt{1 - \widebar{\alpha}_t} \bm{\epsilon}, t) \|^2 \Big] \\
		f_{\phi}(\cdot) & = f_{\phi}(g_{\xi}(\bm{\eta}, \sqrt{\widebar{\alpha}_t} \mathbf{x}_0 + \sqrt{1 - \widebar{\alpha}_t} \bm{\epsilon}, t), t)
	\end{aligned}\right.,
\end{equation} 
where $C_t$ is a constant, $\bm{\eta}, \bm{\epsilon} \sim \mathcal{N}(\mathbf{0}, \mathbf{I})$, and parameters $\theta, \phi, \xi$ are learnable.

For the negative log-likelihood $\mathcal{L}_0^{\mathrm{SMD}} = \mathbb{E}_q [-\ln p^{\mathrm{SMD}} _{\widebar{\theta}} (\mathbf{x}_0 \mid \mathbf{x}_1)  ]$, we expand it as
\begin{equation}
	\mathcal{L}_0^{\mathrm{SMD}}  =  \mathbb{E}_{\mathbf{x}_0, \mathbf{x}_1 \sim q(\mathbf{x}_0)q(\mathbf{x}_1 \mid \mathbf{x}_0)} \Big[ -\ln \Big( \int_{\mathbf{z}_{1}} p_{\widebar{\theta}}^{\mathrm{SMD}}(\mathbf{z}_1 \mid \mathbf{x}_1) p_{\widebar{\theta}}^{\mathrm{SMD}}(\mathbf{x}_0 \mid \mathbf{x}_1, \mathbf{z}_1) d\mathbf{z}_1 \Big) \Big].
\end{equation}
By applying Jensen's inequality, we have
\begin{equation}
	\begin{aligned}
		\mathcal{L}_0^{\mathrm{SMD}} & \le \mathbb{E}_{\mathbf{x}_0, \mathbf{x}_1} \Big[ - \int_{\mathbf{z}_{1}} p_{\widebar{\theta}}^{\mathrm{SMD}}(\mathbf{z}_1 \mid  \mathbf{x}_1) \ln p_{\widebar{\theta}}^{\mathrm{SMD}}(\mathbf{x}_0 \mid \mathbf{x}_1, \mathbf{z}_1) d\mathbf{z}_1 \Big] \\
		& = \mathbb{E}_{\mathbf{x}_0, \mathbf{x}_1} \Big[ \mathbb{E}_{\mathbf{z}_1 \sim p_{\widebar{\theta}}^{\mathrm{SMD}}(\mathbf{z}_1 \mid  \mathbf{x}_1)} [- \ln p_{\widebar{\theta}}(\mathbf{x}_0 \mid \mathbf{x}_1, \mathbf{z}_1) ] \Big] \\
		& = C_1 + \mathbb{E}_{\mathbf{z}_1 \sim p_{\widebar{\theta}}^{\mathrm{SMD}}(\mathbf{z}_1 \mid  \mathbf{x}_1)} \Big[  \mathbb{E}_{\mathbf{x}_0, \mathbf{x}_1} \Big[ \frac{1}{2\sigma_1}\| \mathbf{x}_0 - \bm{\mu}_{\theta, f_{\phi}(\mathbf{z}_1, t)}(\mathbf{x}_1, 1) \|^2 \Big] \Big]
	\end{aligned},
\end{equation}
where $C_1$ is a constant that does not involve with the model parameter $\widebar{\theta} = \theta \bigcup \phi \bigcup \xi$. Considering Eq.~(\ref{eq:forward def}) and Eq.~(\ref{eq:new mean parameterization}), we can convert this inequality into
\begin{equation}
	\label{eq:new loss part-2}
	\begin{aligned}
		\mathcal{L}_0^{\mathrm{SMD}} & \le C_1 + \mathbb{E}_{\mathbf{z}_1 \sim p_{\widebar{\theta}}^{\mathrm{SMD}}(\mathbf{z}_1 \mid  \mathbf{x}_1)} \Big[  \mathbb{E}_{\mathbf{x}_0, \bm{\epsilon}} \Big[  \frac{\beta_1^2}{2\sigma_1\alpha_1(1 - \widebar{\alpha}_1)} \| \bm{\epsilon} - \bm{\epsilon}_{\theta, f_{\phi}(\mathbf{z}_1, t)}(\mathbf{x}_1, 1) \|^2 \Big] \Big] \\
		& = C_1 + \mathbb{E}_{\bm{\eta}, \bm{\epsilon}, \mathbf{x}_0} \Big[  \frac{\beta_1^2}{2\sigma_1\alpha_1(1 - \widebar{\alpha}_1)} \| \bm{\epsilon} - \bm{\epsilon}_{\theta,  f_{\phi}(g_{\xi}(\cdot), t)}(\sqrt{\widebar{\alpha}_1} \mathbf{x}_0 + \sqrt{1 - \widebar{\alpha}_1} \bm{\epsilon}, 1) \|^2  \Big]
	\end{aligned},
\end{equation}
where $\bm{\eta}, \bm{\epsilon} \sim \mathcal{N}(\mathbf{0}, \mathbf{I})$, and the second equality is also derived by LOTUS.

Finally, by combining Eq.~(\ref{eq:new loss part-1}) and Eq.~(\ref{eq:new loss part-2}), we have
\begin{equation}
	\begin{aligned}
		& \mathbb{E}_q[-\log p_{\widebar{\theta}}^{\mathrm{SMD}}(\mathbf{x}_0)]  \le \mathcal{L}^{\mathrm{SMD}} = \sum_{t = 0}^{T} \mathcal{L}^{\mathrm{SMD}}_t \\
		& = C + \sum_{t=1}^{T}  \mathbb{E}_{\bm{\eta}, \bm{\epsilon}, \mathbf{x}_0} \Big[ \Gamma_t \| \bm{\epsilon} - \bm{\epsilon}_{\theta, f_{\phi}(\cdot)} (\sqrt{\widebar{\alpha}_t} \mathbf{x}_0 + \sqrt{1 - \widebar{\alpha}_t} \bm{\epsilon}, t)  \|^2 \Big]
	\end{aligned},
\end{equation}
where $C = \sum_{t = 1}^T C_t$ and $\Gamma_t = \beta_t^2 / (2\sigma_t\alpha_t(1 - \widebar{\alpha}_t))$.

\section{Generated Samples}

\begin{figure}
		\centering
		\begin{subfigure}{0.48\textwidth}
			\includegraphics[width=\linewidth]{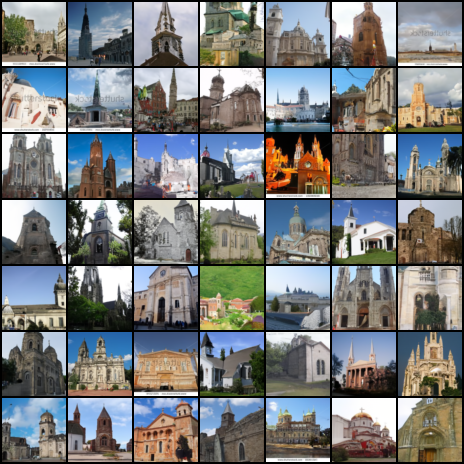}
			\caption{Synthesized images of LSUN Church}
		\end{subfigure}
		\hfill
		\begin{subfigure}{0.48\textwidth}
			\includegraphics[width=\linewidth]{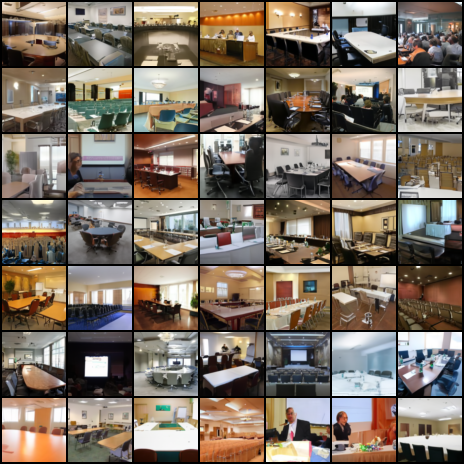}
			\caption{Synthesized images of LSUN Conference}
		\end{subfigure}
		\caption{$64 \times 64$ images generated by DDPM w/ SMD.}
		\label{fig:sampled images 1}
	\end{figure}

\begin{figure}
		\centering
		\begin{subfigure}{0.715\textwidth}
			\includegraphics[width=\linewidth]{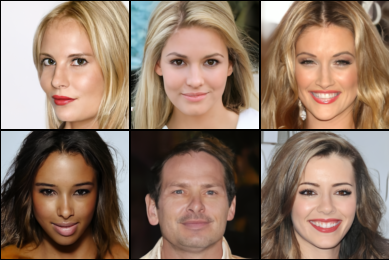}
		\end{subfigure}
		\hfill
		\begin{subfigure}{0.24\textwidth}
			\includegraphics[width=\linewidth]{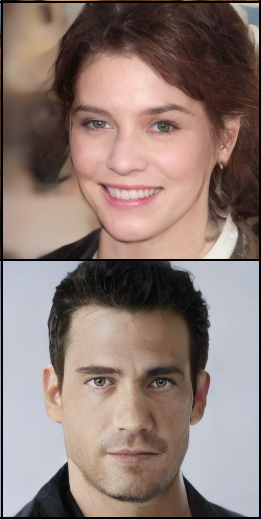}
		\end{subfigure}
		\caption{Generated images on CelebA-HQ $128 \times 128$ (left) and $256 \times 256$ (right). The left samples are from DDPM w/ SMD and the right ones from LDM w/ SMD.}
		\label{fig:sampled images 2}
	\end{figure}

	Some images generated by our models (e.g., LDM w/ SMD) are in Fig.~\ref{fig:sampled images 1} and Fig.~\ref{fig:sampled images 2}. 

\end{document}